\documentclass[sigconf]{acmart}
\AtBeginDocument{%
  \providecommand\BibTeX{{%
    \normalfont B\kern-0.5em{\scshape i\kern-0.25em b}\kern-0.8em\TeX}}}

\setcopyright{none}
\renewcommand\footnotetextcopyrightpermission[1]{}

\usepackage{amsmath,amssymb,amsfonts}
\usepackage{algorithmic}
\usepackage{graphicx}
\usepackage{textcomp}
\usepackage{xcolor}
\def\BibTeX{{\rm B\kern-.05em{\sc i\kern-.025em b}\kern-.08em
    T\kern-.1667em\lower.7ex\hbox{E}\kern-.125emX}}

\usepackage[ruled,vlined,linesnumbered]{algorithm2e}
\usepackage{setspace}
\usepackage{xspace}
\usepackage{enumitem}
\usepackage{url}
\usepackage{multirow}
\usepackage{hyperref}
\hypersetup{hidelinks} 
\usepackage{booktabs}
\usepackage{tcolorbox}
\usepackage{bbm}
\usepackage{balance}

\def \projectName {\textsc{MwT}\xspace}   
\def \projectNameCNN {\textsc{MwT}\xspace}   

\begin{document}

\title{Modularizing while Training: A New Paradigm for Modularizing DNN Models}

\author{Binhang Qi}
\affiliation{
  \institution{SKLSDE Lab, Beihang University\country{China}}
}
\email{binhangqi@buaa.edu.cn}

\author{Hailong Sun\textsuperscript{\textdagger}}
\affiliation{
  \institution{SKLSDE Lab, Beihang University\country{China}}
}
\email{sunhl@buaa.edu.cn}

\author{Hongyu Zhang}
\affiliation{
  \institution{Chongqing University
  \\\country{China}}
}
\email{hyzhang@cqu.edu.cn}

\author{Ruobing Zhao}
\affiliation{
  \institution{SKLSDE Lab, Beihang University\country{China}}
}
\email{rbingzhao@buaa.edu.cn}

\author{Xiang Gao\textsuperscript{\textdagger}}
\affiliation{
  \institution{SKLSDE Lab, Beihang University\country{China}}
}
\email{xiang_gao@buaa.edu.cn}

\begin{abstract}
Deep neural network (DNN) models have become increasingly crucial components of intelligent software systems. 
However, training a DNN model is typically expensive in terms of both time and computational resources. 
To address this issue, recent research has  focused on reusing existing DNN models - borrowing the concept of software reuse in software engineering.
However, reusing an entire model could cause extra overhead or inherit the weaknesses from the undesired functionalities.
Hence, existing work proposes to decompose an already trained model into modules, i.e., \textit{modularizing-after-training}, to enable module reuse.
Since the trained models are not built for modularization, modularizing-after-training may incur huge overhead and model accuracy loss.
In this paper, we propose a novel approach that incorporates modularization into the model training process, i.e., \textit{modularizing-while-training} (\projectName).
We train a model to be structurally modular through two loss functions that optimize intra-module cohesion and inter-module coupling. %
We have implemented the proposed approach for modularizing Convolutional Neural Network (CNN) models.
The evaluation results on representative models demonstrate that \projectName outperforms the existing state-of-the-art modularizing-after-training approach. Specifically, the accuracy loss caused by \projectName is only 1.13 percentage points, which is less than that of the existing approach. %
The kernel retention rate of the modules generated by \projectName is only 14.58\%, 
with a reduction of 74.31\% over the existing approach.
Furthermore, the total time cost required for training and modularizing is only 108 minutes, which is half the time required by the existing approach. 
Our work demonstrates that \projectName is a new and more effective paradigm for realizing DNN model modularization, offering a fresh perspective on achieving model reuse.


\end{abstract}



\maketitle

\renewcommand{\thefootnote}{\fnsymbol{footnote}}
\footnotetext[2]{Corresponding authors: Hailong Sun and Xiang Gao. Hailong Sun is also with Hangzhou Innovation Institute, Beihang University, China.}

\pagestyle{plain}

\newcommand{\toolname}[0]{\textsc{MwT}}
\section{Introduction}

Software reuse~\cite{software_reuse_1,software_reuse_2,software_reuse_3} can facilitate software development. According to Gartner\cite{gartner}, over 95\% of companies use at least one open-source software component in building their business products, highlighting the importance of software reuse. Like open-source software, more and more deep neural network (DNN) models, trained for various tasks~\cite{qi2021dreamloc,tao2021towards,yao2019graph}, are also publicly available on model sharing platforms such as GitHub and Hugging Face~\cite{huggingface}. These DNN models have become increasingly crucial components of intelligent software systems. Reusing these models to facilitate the development of intelligent software systems has sparked interests in recent years~\cite{icse21discriminiate,modeldiff,ji2018model,wang2015model,qi2022patching,icse2023,nnmodularity2022icse}.

However, %
reusing a whole DNN model causes extra overhead~\cite{icse2023,qi2022patching} or inherits the weakness from the undesired functionalities~\cite{icse2023,ReMos,liu2018fine}.
Modularization~\cite{parnas1972criteria,parnas1976design,tarr1999n} is a software design technique that involves separating %
a program into independent and reusable %
modules. Each module is self-contained and responsible for executing a part of the desired functionality. This paradigm enables software reuse by allowing developers to reuse specific packages or functions in their programs. Borrowing the idea of modularization in software engineering, researchers propose to modularize DNN models, then reuse the resultant modules. For example, Qi et al.~\cite{qi2022patching,icse2023} and Pan et al.~\cite{fse2020modularity, nnmodularity2022icse} proposed to decompose a trained DNN model into modules by identifying neurons or weights that are responsible for classifying each class and removing irrelevant neurons and weights.
They then use the resultant modules for constructing, patching, or transferring DNN models. %

Existing work~\cite{qi2022patching,icse2023,nnmodularity2022icse,fse2020modularity,ReMos,deeparc} on modularizing DNN models follows the paradigm of \textit{modularizing after training}, i.e., they first train a model as a whole or find an already trained model, then decompose the model into modules.
However, since the model is not directly trained for modularization, decomposing trained models %
into modules
has three main limitations~\cite{qi2022patching, nnmodularity2022icse, icse2023}.
First, the obtained modules may share a large portion of weights, meaning that the weights relevant to implementing different functionalities have large overlaps.
When decomposing a trained model for a certain prediction task, the corresponding modules may retain a large number of weights.
For instance, modules generated by ~\cite{fse2020modularity} retain 76.21\% of the weights.
CNNSplitter~\cite{qi2022patching} achieves better performance, but its resulting modules still retain 56.76\% of the convolution kernels.
Second, decomposing trained models could cause huge resource and time consumption.
For instance, CNNSplitter costs at least 3 hours to decompose a simple trained model with 4,224 convolution kernels.
Last but not least, the modularization process may cause non-trivial accuracy loss, e.g., the average loss of accuracy caused by CNNSplitter is 2.89\%.
Those limitations affect the practical usability of the paradigm of \textit{modularizing after training}.

To overcome the above limitations, in this paper, we propose \textit{Modularizing while Training} (\projectName), a new paradigm for modularizing DNN models. %
It is well known that modular software development requires (1) \textit{high cohesion}, which means keeping code that is closely related to each other in a single module, and (2) \textit{low coupling}, which means separating unrelated code into different modules~\cite{booch2008object}.
Inspired by the principle of modular software development, \projectName trains models with the goal of achieving \textit{higher cohesion} and \textit{lower coupling} as much as possible in the \textit{modular training} stage.
In our paradigm, high cohesion means keeping the weights relevant to a certain prediction task into a small portion of weights of the whole model, while low coupling means the weights used for different prediction tasks have small overlaps.
Given a DNN model that is trained with high cohesion and low coupling, the \textit{modularizing} stage can achieve more efficient and effective decomposition than the existing approach of modularizing after training~\cite{fse2020modularity,qi2022patching,nnmodularity2022icse}. %

To realize this idea, we first define cohesion and coupling in the context of DNN modularization. \projectName measures cohesion by computing the overlap between weights corresponding to the samples belonging to the same class. \projectName measures the coupling by computing the overlap between weights corresponding to different modules. 
To train high-cohesion and low-coupling models, %
we design cohesion loss and coupling loss functions and integrate them with the cross-entropy loss.
Then to decompose a trained model, %
\projectName simply needs to remove the irrelevant weights to construct modules. %
Based on the \projectName framework, we design and implement a concrete approach for CNN modularization. The reason for choosing CNN is that the CNN model is a mainstream DNN model that has been the focus of existing DNN modularization work.

We evaluate \projectName using four representative CNN models on two widely-used datasets. 
The experimental results demonstrate that \projectName can maintain sufficient model classification accuracy while effectively improving cohesion and reducing coupling. Compared to the existing state-of-the-art modularizing-after-training approach~\cite{qi2022patching}, the loss of accuracy caused by \projectName is only 1.13 percentage points, which is 
1.76 percentage points less than that of the existing approach. 
Notably, the kernel retention rate of modules generated by \projectName is only 14.58\%, 
with a reduction of 74.31\% over the existing approach. The total time cost of training and modularizing is 108 minutes, which is half the time required by the existing approach. 
Moreover, module reuse leads to significantly lower inference cost than reusing the entire model.

The main contributions of this work are as follows:

\begin{itemize}[leftmargin=*]
    \item We propose a new paradigm for modularizing DNN models, called \textit{modularizing while training}, which achieves more efficient and effective decomposition than the current paradigm of  
    \textit{modularizing after training}. 
    \item 
    We propose a framework called \projectName and implement a concrete tool for CNN modularization based on \projectName. 
    We propose strategies for recognizing relevant kernels, and then design loss functions for evaluating cohesion and coupling.
    \item We conduct extensive experiments on two widely-used datasets using four representative CNN models. The results demonstrate that \projectName can maintain the model's classification accuracy while improving cohesion and reducing coupling. Moreover, our experiments show that \projectName outperforms the state-of-the-art approach in terms of both effectiveness and efficiency.

\end{itemize}

\section{Preliminaries}
\subsection{Mainstream neural networks}
Neural networks (NNs)~\cite{alexnet,huang2017densely,graves2014towards} are a type of machine learning models that comprise interconnected layers of neurons. The connections between neurons are represented by weights, which are the essential structure and learnable parameters that determine the functionality and performance of the models. The most basic NN is the fully connected neural network, in which each neuron in one layer is connected to all neurons in the next layer.

With the advancement of neural networks, various network structures have been developed for processing different types of data. Among them, CNN~\cite{lenet, googlenet, szegedy2017inception} is a popular type of NN specifically designed for image data and has been widely adopted in computer vision~\cite{pan2022integration,xie2021oriented} and various software engineering tasks~\cite{wu2022vulcnn, lee2022light,sun2019grammar,huo2019deep}. A CNN model typically consists of convolutional layers, pooling layers, and fully connected layers, with convolutional layers being the core of CNNs~\cite{cnn2018overview,alexnet}. Each convolutional layer contains numerous convolution kernels, which consists of a group of weights. Each kernel learns to recognize local features of an input tensor and outputs a feature map that reflects the degree of matching between the kernel and the input tensor.

\subsection{Problem Formulation}
Given a trained $N$-class classification model $\mathcal{M}=(\mathcal{N}, \mathcal{W})$, where $\mathcal{N}$ and $\mathcal{W}$ are the sets of neurons and weights in the model, respectively, and a training sample set of $\mathcal{M}$, denoted as $\mathcal{S}=\{\mathcal{S}_n\}_{n=1}^{N}$, where $\mathcal{S}_n$ represents the set of samples belonging to class $n$, the formal definition of DNN modularization is as follows:
\vspace{5pt}

\textsc{Definition} 1. \textit{\textbf{DNN Modularization} aims to compute a set of modules, denoted as $\{m_n\}_{n=1}^N$. To compute $m_n$, modularization recognizes subsets $\mathcal{N}_n \subset \mathcal{N}$ and $\mathcal{W}_n \subset \mathcal{W}$ used to classify $\mathcal{S}_n$.}
\vspace{5pt}

The set of weights $\mathcal{W}_n$ can consist of individual weights or substructures of NNs. Specifically, since weights are fundamental structures of NNs, weight sets based on individual weights are applicable to all DNN models for modularization.
In addition, weight sets based on substructures such as convolution kernels for CNNs or attention heads for Transformers are applicable to the DNN models constructed mainly with the corresponding NNs for modularization.
To evaluate modularization, the definitions of cohesion and coupling in the context of DNN modularization are as follows:

\vspace{5pt}

\textsc{Definition} 2. \textit{\textbf{Cohesion} of module $m_n$ is the degree of overlap between the sets of weights $\{\mathcal{W}_n^i\}_{i=1}^{|\mathcal{S}_n|}$, where $\mathcal{W}_n^i$ represents the set of weights used to classify the $i$-th sample in $\mathcal{S}_n$.}
\vspace{5pt}

\textsc{Definition} 3. \textit{\textbf{Coupling} between two modules $m_n$ and $m_k$ is the degree of overlap between sets of weights $\mathcal{W}_n$ and $\mathcal{W}_k$.}
\vspace{5pt}

In the end, the cohesion and coupling of the result of modularization are calculated as the average cohesion of all modules and the average coupling across all pairs of modules, respectively.

\section{Modularizing while Training}
\label{sec:approach}


In this section, we present the detailed methodology of Modularizing-While-Training (MwT).
Specifically, Section \ref{subsec:overview} introduces the general framework that incorporates modularization into the model training process. A concrete approach for modularizing convolutional neural networks using \projectName is then presented in Section \ref{subsec:modulartraining} and Section \ref{subsec:modularizing}. Additionally, Section \ref{subsec:reuse} introduces the concept of on-demand model reuse based on \projectName.

\subsection{Overview}
\label{subsec:overview}

\begin{figure}
    \centering
    \includegraphics[width=\columnwidth]{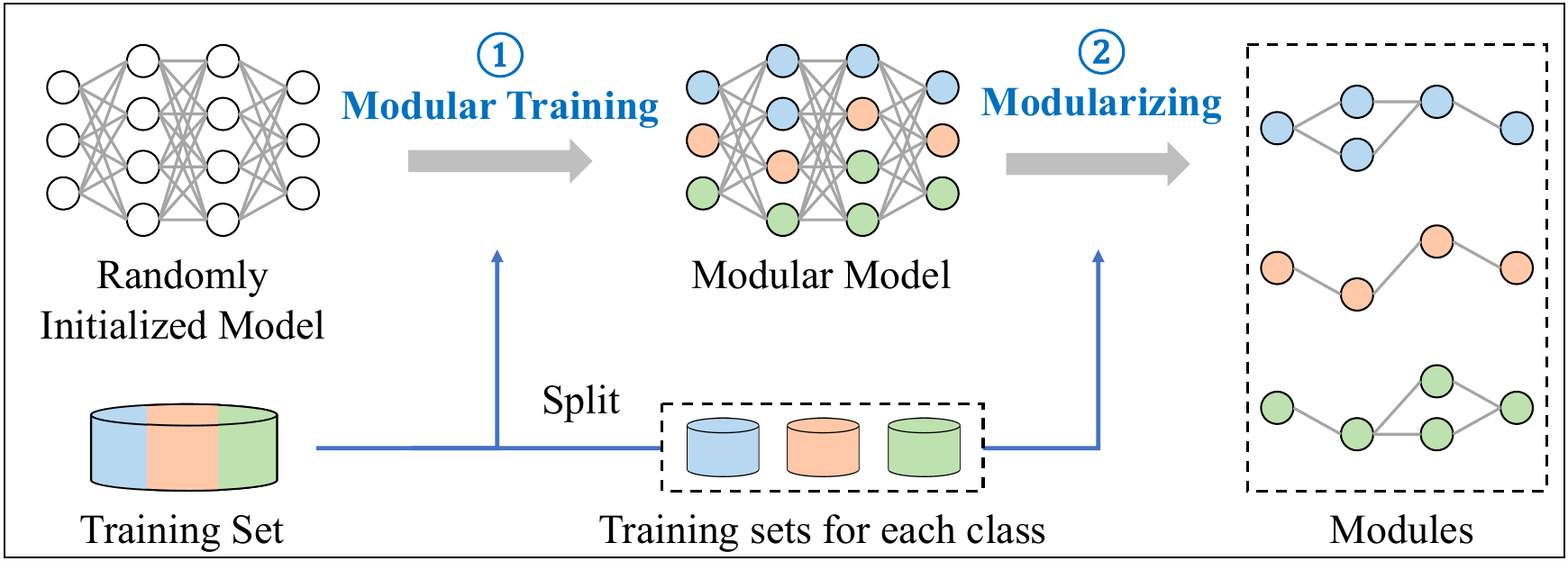}
    \caption{An overview of the proposed framework \textit{modularizing while training}.}
    \label{fig:framework}
\end{figure}

The goal of DNN modularization is to decompose a model into modules, each corresponding to a class and containing only the weights necessary for classifying samples of that class. To achieve this goal and overcome the limitations of \textit{modularizing after training} approaches, we design a novel \textit{modularizing while training} framework, called \projectName, as shown in Figure \ref{fig:framework}.
From a high-level perspective, \projectName consists of two stages: \textit{modular training} and \textit{modularizing}.

\begin{figure*}
    \centering
    \includegraphics[width=\textwidth]{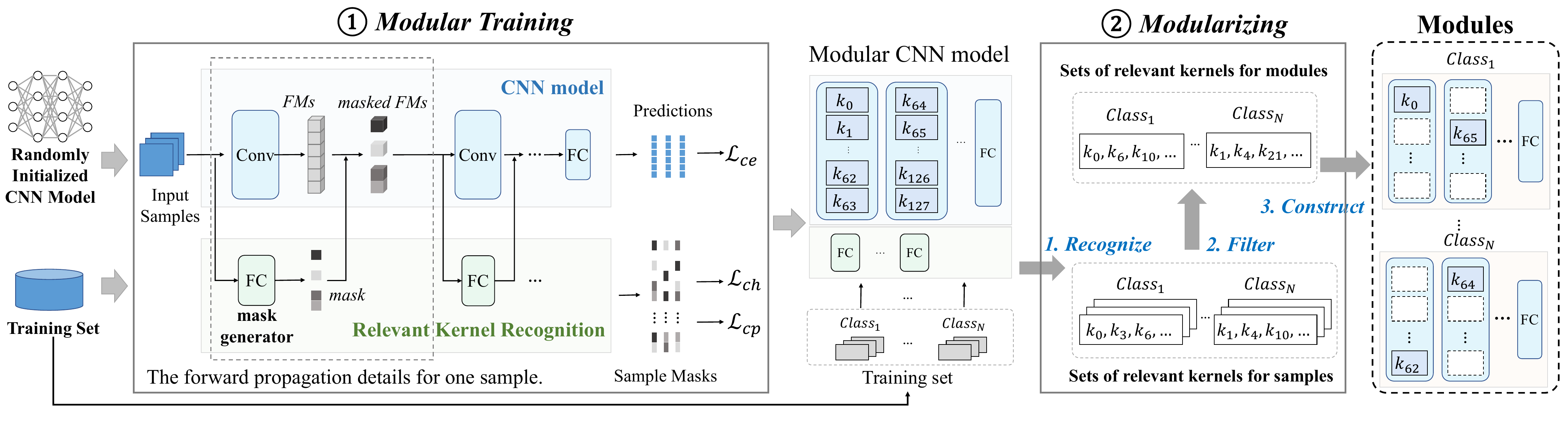}
    \vspace{-18pt}
    \caption{The workflow of \projectName for CNN models.}
    \label{fig:mwt4cnn}
\end{figure*}

The \textit{modular training} stage is responsible for training %
a modular model that has sufficient classification ability and performs well in terms of modularity metrics (i.e., coupling and cohesion).
Enhancing classification accuracy while improving model performance in coupling and cohesion during training is the core of modular training. 
During modular training, \projectName recognizes the weights required to classify each training sample. Then, it evaluates the cohesion of the modular model by computing the overlap between weights corresponding to training samples belonging to the same class.
\projectName evaluates the coupling by computing the overlap between weights corresponding to training samples belonging to different classes.
Based on the computation of coupling and cohesion, \projectName designs coupling loss and cohesion loss functions. While minimizing the classification loss (e.g., cross-entropy loss) using gradient descent, \projectName also minimizes the coupling loss and cohesion loss to optimize the performance of the modular model in terms of coupling and cohesion.
Through iterative optimization, the resulting modular model has sufficient classification capability and achieves high cohesion and low coupling in terms of modularity.

The \textit{modularizing} stage decomposes the resulting modular model into modules using the training samples of each class. Each module retains only the weights responsible for the corresponding class.

In this work, we apply \projectName to CNN models. CNN is a mainstream neural network model, whose modularization has been the focus of existing works~\cite{qi2022patching, nnmodularity2022icse, icse2023}. 
Following existing modularization work, we propose a concrete approach for CNN modularization based on \projectName. %
It is worth mentioning that, inspired by CNNSplitter~\cite{qi2022patching}, \projectNameCNN generates modules at the granularity of convolution kernels rather than individual weights. 
In this way, the generated module contains fewer weights than the model, and the module can perform well without the requirement of special libraries~\cite{qi2022patching,icse2023}.

\subsection{Modular Training}
\label{subsec:modulartraining}

Training a modular CNN model involves three phases: (1) recognition of relevant convolution kernels, (2) evaluation of the cohesion and coupling, and (3) optimization of the modular model's performance with regard to cohesion and coupling. 

\subsubsection{Recognition of relevant convolution kernels}
As illustrated in Figure \ref{fig:mwt4cnn}, to recognize the relevant kernels for an input sample, a \textit{relevant kernel recognition} is appended to the original CNN model and trained jointly with the model.
The relevant kernel recognition consists of \textit{mask generators}, each of which is a fully connected (FC) layer and corresponds to a convolutional layer, thereby enabling it to recognize the kernels responsible for the input sample based on the convolutional layer's input. 
Taking an example of a convolutional layer in the dashed box in Figure \ref{fig:mwt4cnn}, 
the input sample to the convolutional layer, denoted as $input$, is an image or the output of the previous convolutional layer.
Suppose $input$ has a dimension of $(C, H, W)$, where $C$ represents the number of channels, $H$ denotes the height of input planes in pixels, and $W$ indicates the width in pixels.
First, $input$ is fed simultaneously to the convolutional layer $Conv$ and the mask generator.
$Conv$ comprises $N_k$ kernels, which produce $N_k$ feature maps, denoted as $FMs$.
Meanwhile, the mask generator outputs a mask with a dimension of $N_k$ based on $input$.
The $N_k$ elements of the mask correspond to the $N_k$ convolution kernels, representing whether each kernel is relevant to $input$.
The activation functions of the mask generator are $Tanh$ and $ReLU$, which map the value of its outputs to $[0, 1)$, resulting in the mask.
The kernels associated with the elements in the mask having values greater than 0 are relevant. 
Specifically, $FMs$ are multiplied by the mask, which could filter out the feature maps produced by irrelevant kernels.
Based on the produced masks, \projectName recognizes the relevant kernels for each sample.

Besides, to reduce the overhead for training relevant kernel recognition, before being fed to the mask generator, $input$ is subject to an average pooling operation to obtain a tensor $input'$ with a smaller dimension of $C$. 
Reducing the dimension of input could significantly lower the computational cost.

\subsubsection{Calculation of cohesion and coupling}
Using the generated sample masks as described above, we can obtain the set of relevant kernels for each sample.
Specifically, for a certain class $c_i$, the $n_i$ samples belonging to $c_i$ are denoted as \{$s_i^1$, $s_i^2$, $\dots$ $s_i^{n_i}$\}, and their corresponding kernel sets are represented as \{$sK_i^1$, $sK_i^2$, $\dots$ $sK_i^{n_i}$\}.
The kernel set $mK_i$ constituting module $m_i$ is calculated as $mK_i=\bigcup_{k=1}^{n_i} sK_i^k$, meaning that if one kernel is useful for any sample of class $c_i$, it will be retained in the corresponding module $m_i$.
\projectNameCNN evaluates the cohesion of $m_i$ by measuring the similarity between the kernel sets $sK_i^j$ and $sK_i^k$ for $0<j<k\leq n_i$. %
The similarity metrics are defined as the Jaccard Index ($JI$), which is also utilized by existing work~\cite{qi2022patching,nnmodularity2022icse} to measure the similarity between modules.
Specifically, the $JI$ between sets $A$ and $B$ is obtained by dividing the size of the intersection of two sets by the size of the union:
\begin{gather}
    JI(A, B) = \frac{|A \cap B|}{|A \cup B|}. \label{eq:jaccard}
\end{gather}
$JI(A, B) = 1$ indicates that the two sets are exactly the same; conversely, a value of 0 indicates that there is no overlap between the two sets.
Based on $JI$, the cohesion of the module $m_i$ responsible for the class $c_i$ is calculated as follows:
\begin{gather}
    Cohesion(m_i) = \frac{2}{n_i \times (n_i-1)} \times \sum_{0<j<k\leq n_i} JI(sK_i^j, sK_i^k). \label{eq:cohesion}
\end{gather}

In other words, the cohesion of module $m_i$ is calculated by the averaged $JI$ of each pair of $sK_i^j$ and $sK_i^k$.
On the other hand, \projectNameCNN evaluates the coupling between $m_i$ and $m_j$ by measuring the overlap between their relevant kernel sets $mK_i$ and $mK_j$.
Specifically, the coupling between $m_i$ and $m_j$ is calculated as follows:
\begin{gather}
    Coupling(m_i, m_j) = JI(mK_i, mK_j). \label{eq:coupling}
\end{gather}
In the end, the cohesion and coupling of the modular model are calculated as the average cohesion of all modules and the average coupling across all pairs of modules, respectively.

\subsubsection{Modular training to optimize accuracy, cohesion, and coupling}
To continuously improve the performance of the modular model in terms of cohesion and coupling during training, a straightforward way is to integrate the two metrics into the loss function. 
Then, the optimization goal is to boost the modular model's classification accuracy while improving its cohesion and reducing its coupling.
Through gradient descent, the cohesion and coupling losses, as well as the cross-entropy loss, are minimized to optimize the CNN model and relevant kernel recognition.
Common gradient descent methods can only perform optimization in continuous space, but convolution kernel sets are discrete,
which makes it hard to use the Jaccard Index-based cohesion and coupling as the loss function.
Therefore, we calculate the cohesion and coupling losses by computing the cosine similarity between the masks of samples. 
Specifically, given a batch of training samples, the cohesion loss $\mathcal{L}_{ch}$ and coupling loss $\mathcal{L}_{cp}$ are computed as follows:
\begin{small}
\begin{gather}
    \mathcal{L}_{ch} = 1 - \frac{1}{N} \times \sum_{i=1}^{N} \left(\sum_{0 < j < k \leq n_i} Cos(sM_i^j, sM_i^k) \div \frac{n_i \times (n_i-1)}{2} \right), \label{eq:cohesion_loss} \\
    \mathcal{L}_{cp} = \frac{2}{N^2-N} \times \sum_{0 < h < i \leq N} \left(\frac{1}{n_h \times n_i} \times \sum_{j=1}^{n_h}\sum_{k=1}^{n_i} Cos(sM_h^j, sM_i^k)\right), \label{eq:coupling_loss}
\end{gather}
\end{small}
where $N$ is the number of classes, $n_i$ is the number of samples in the batch belonging to the class $c_i$, $sM_i^j$ represents the mask of the $j$-th sample of class $c_i$, 
and $Cos(*, *)$ is the cosine similarity between two tensors.
Since the range of element values in the mask is $[0,1)$, the cosine similarity between two masks is within $[0,1]$. Thus the ranges of both $\mathcal{L}_{ch}$ and $\mathcal{L}_{cp}$ are $[0, 1]$.
With all loss functions, the objective function is defined as follows:
\begin{gather}
    \mathcal{O}=\mathcal{L}_{ce} + \alpha \times \mathcal{L}_{ch} + \beta \times \mathcal{L}_{cp}, \label{eq:objective}
\end{gather}
where $\mathcal{L}_{ce}$ is the cross-entropy loss, $\alpha$ and $\beta$ are weighting factors.

To minimize $\mathcal{O}$, modular training simultaneously minimizes $\mathcal{L}_{ce}$, $\mathcal{L}_{ch}$, and $\mathcal{L}_{cp}$ through mini-batch gradient descent. By minimizing $\mathcal{L}_{ce}$, the modular model learns to use the corresponding kernels (i.e., modules) to classify different classes of samples.
By minimizing $\mathcal{L}_{ch}$, the modular model tends to generate identical masks for samples belonging to the same class, i.e., improving the cohesion of the modular model.
By minimizing $\mathcal{L}_{cp}$, the modular model tends to generate masks with zero cosine similarity for samples belonging to different classes.
Therefore, minimizing the coupling loss can reduce the coupling of the modular model. Modular training iterates specified epochs and outputs the modular model.

\subsection{Modularizing}
\label{subsec:modularizing}

During modular training, a modular model including the CNN model and relevant kernel recognition is generated.
To decompose the trained modular model into modules, \projectName first generates masks for each module using the trained relevant kernel recognition.
To generate a mask for module $m_i$, the basic idea is (1) for all training sample \{$s_i^1$, $s_i^2, \dots s_i^{n_i}$\} belonging to class $c_i$, using relevant kernel recognition to generate sets of sample masks \{$sM_i^1$, $sM_i^2, \dots, sM_i^{n_i}$\}, and (2) calculate the module's mask as 
$mM_i = \{\frac{\mathit{kernel\_count}}{n_i} > \tau : 1, 0\ |\ \mathit{kernel\_count} \in \sum_{j=1}^{n_i} sign(sM_i^j) \}$, meaning that if the frequency of a kernel occurring in samples' kernel sets exceeds a threshold $\tau$, the element in the module's mask corresponding to the kernel will be marked as $1$, otherwise, it will be marked as $0$.
The kernels corresponding to the elements of $mM_i$ marked as $1$ are relevant to $m_i$.
\projectName filters out kernels by setting a reasonable threshold $\tau$ because of the inherent randomness of neural network models.
It is challenging for the mask generator to ensure identical masks for different samples of the same class. 
Especially when the number of training samples is large, it is inevitable that some individual relevant kernels of a sample are irrelevant to other samples. 
Simply regarding all relevant kernels of samples as module's kernel may result in a module containing numerous redundant kernels.

\begin{figure}
    \centering
    \includegraphics[width=0.8\columnwidth]{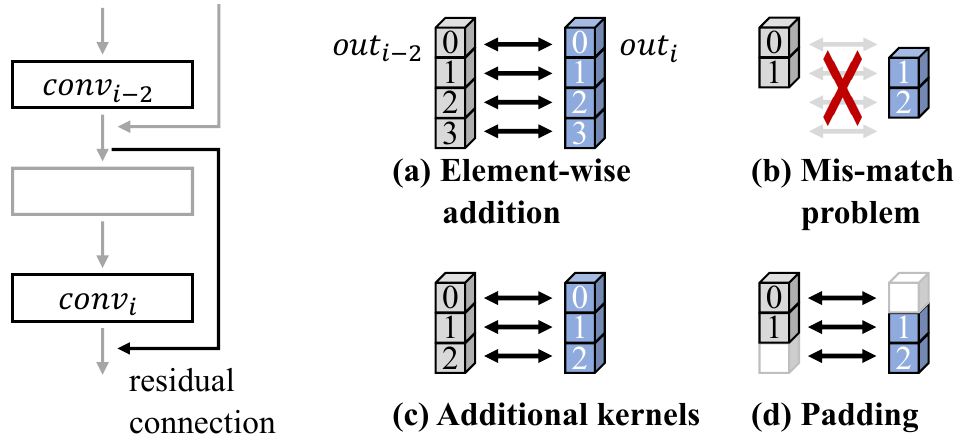}
    \vspace{-6pt}
    \caption{The illustration of residual connections and dimension mismatch problem.}
    \label{fig:remove_kernels}
\end{figure}

After obtaining the relevant kernels of a module, irrelevant kernels in the CNN model are removed, and the relevant kernel recognition is eliminated, resulting in a module responsible for class $c_i$. 
In simple CNN models, such as VGG~\cite{vgg}, which consist of stacked convolutional layers, the removal of irrelevant kernels is straightforward.
Specifically, when removing irrelevant kernels, the retained kernels' channels corresponding to the irrelevant kernels in the previous convolutional layer are also eliminated.
However, for complex CNN models containing residual connections (such as ResNet~\cite{resnet}), removing irrelevant kernels may %
cause two layers connected with a residual connection to produce mismatched outputs.
Figure~\ref{fig:remove_kernels} presents a residual network, where layers $conv_i$ and $conv_{i-2}$ have a residual connection.
In the residual network, the outputs $out_i$ and $out_{i-2}$ of $conv_i$ and $conv_{i-2}$ are element-wise added, as shown in Figure \ref{fig:remove_kernels}-(a).
As shown in Figure \ref{fig:remove_kernels}-(b), when the irrelevant kernels are removed from $conv_i$ and $conv_{i-2}$, their outputs may mismatch in the dimension of channel, resulting in failed element-wise addition.
To address this problem, existing work CNNSplitter retains more kernels to keep $out_i$ and $out_{i-2}$ consistent.
More specifically, in addition to retaining their respective relevant kernels, $conv_i$ and $conv_{i-2}$ must also retain extra irrelevant kernels to ensure that both $out_i$ and $out_{i-2}$ have the same number of channels, as depicted in Figure \ref{fig:remove_kernels}-(c).
Nevertheless, these additional kernels result in extra convolution computation, which could lead to much overhead. 
To avoid this issue, we design a padding-based method to address the mismatch problem. 
As illustrated in Figure \ref{fig:remove_kernels}-(d), $conv_i$ and $conv_{i-2}$ each retain their relevant kernels.
Before the element-wise addition of $out_i$ and $out_{i-2}$, zero padding is applied to them to ensure that they have the same number of channels. %
Since the computation of the padding operation is extremely low, the overhead of our padding-based method is lower than that of the existing method.

\begin{figure}
    \centering
    \includegraphics[width=0.85\columnwidth]{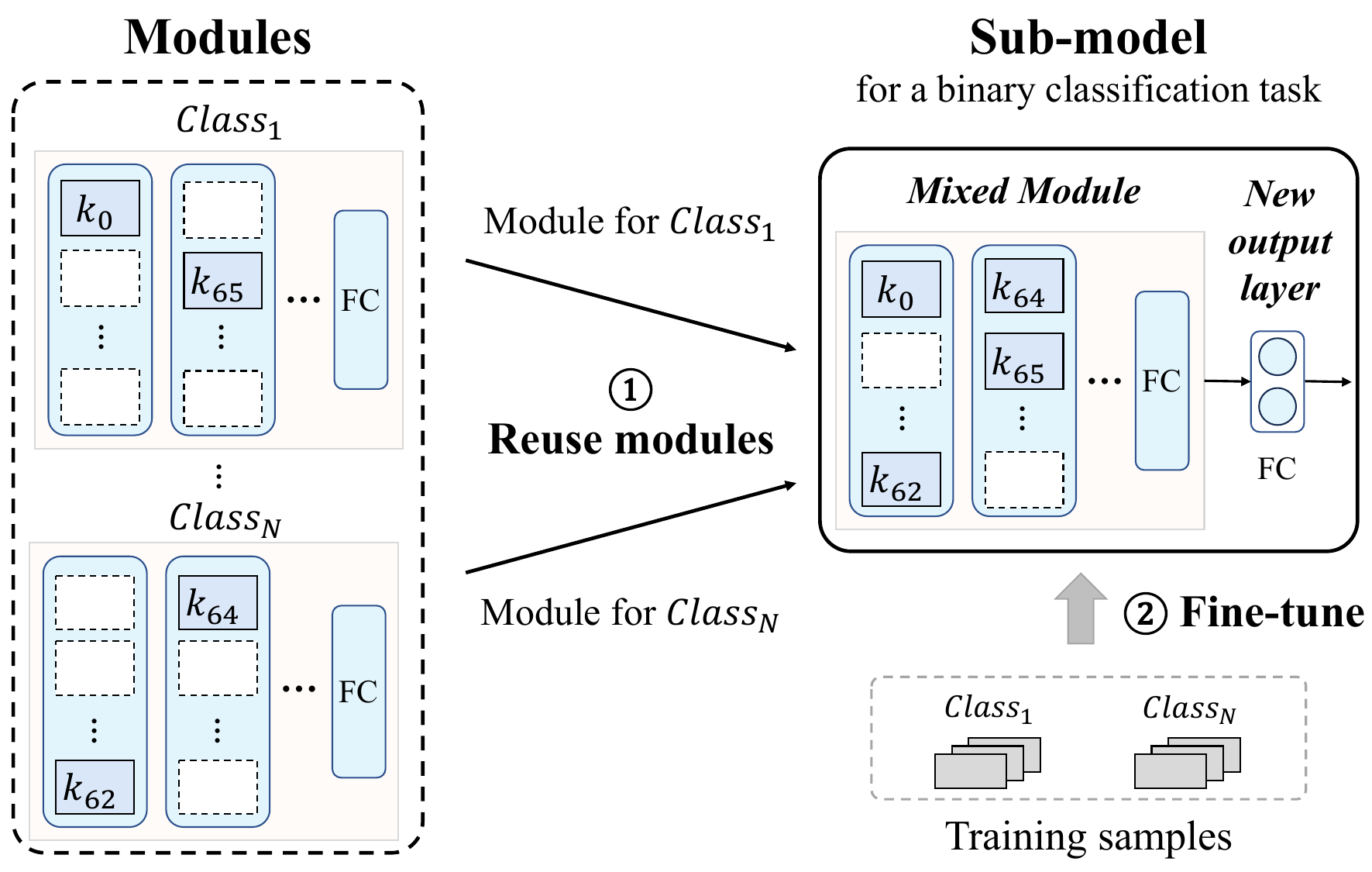}
    \vspace{-6pt}
    \caption{The illustration of on-demand model reuse.}
    \label{fig:module_reuse}
\end{figure}

\subsection{On-demand Model Reuse}
\label{subsec:reuse}
\projectName can better achieve on-demand model reuse, which means dynamically selecting modules (rather than the whole model) from 
the model
and composing them according to specific user requirements. 
As users usually do not need all the classification functionality, on-demand model reuse %
can construct a sub-model with fewer weights than the original model, resulting in lower inference costs. 

Specifically, as illustrated in Figure \ref{fig:module_reuse}, to reuse the $N$-class model on demand on a sub-task containing $M$ classes (i.e., $Class_1$ and $Class_N$ in the example), the $M$ modules responsible for these $M$ classes are reused. 
When the reused modules originate from the same trained model, \projectName constructs a \textit{mixed module} based on the union of convolution kernel sets of these modules, in contrast to the existing work~\cite{fse2020modularity,nnmodularity2022icse} that constructs a composed model by treating each module as an individual classifier. As different modules could contain some of the same weights, a mixed module can avoid redundant retention of the same weights compared to a composed model.
Since the output layer of the mixed module still corresponds to the $N$ classes of the 
original task, a randomly initialized FC layer with a dimension of $(N, M)$ is appended as a \textit{new output layer} after the mixed module, mapping the $N$-dimension output to an $M$-dimension output.
As a result, an $M$-class classification sub-model is constructed, which contains only the kernels responsible for the $M$ classes. The sub-model only needs to be fine-tuned for a few epochs (e.g., 5 epochs) using samples belonging to these $M$ classes from the training set of the original model, thus achieving comparable accuracy to the original model on the target task. Note that the fine-tuning of the sub-model does not use additional data beyond the training samples of the original model.%

When reused modules come from different trained models, mixed modules can be separately constructed based on each model's modules. Subsequently, similar to existing work~\cite{nnmodularity2022icse}, a composed model could be constructed by combining these mixed modules and be fine-tuned. We will explore this reuse scenario in future work.

\section{Experiments}
\label{sec:exp}
To verify the effectiveness of \projectNameCNN,  we present the benchmarks and experimental setup as well as the experimental results.
Specifically, we evaluate \projectNameCNN by answering the following research questions:
\begin{itemize}[leftmargin=*]
    \item RQ1: How effective is \projectNameCNN in training and modularizing CNN models?
    \item RQ2: How efficient is \projectNameCNN in training and modularizing CNN models?
    \item RQ3: How effective is \projectNameCNN in reusing CNN modules?
    \item RQ4: How do the major hyper-parameters influence the performance of \projectNameCNN?
\end{itemize}

\subsection{Experimental Setup}
\label{subsec:setup}
\textbf{Models.} Two representative CNN models including ResNet18~\cite{resnet} and VGG16~\cite{vgg}, and two CNN models, SimCNN and ResCNN from the baseline~\cite{qi2022patching} are utilized to evaluate the effectiveness of \projectNameCNN. 

\textbf{Datasets.} Two public classification datasets are used to train and modularize the CNN models, including CIFAR10~\cite{cifar10} and Street View House Number (SVHN)~\cite{svhn}, which are also used in the baseline approaches~\cite{qi2022patching,nnmodularity2022icse}.
The CIFAR10~\cite{cifar10} dataset contains natural images with resolution 32 × 32, which are drawn from 10 classes including \emph{airplanes}, \emph{cars}, \emph{birds}, \emph{cats}, \emph{deer}, \emph{dogs}, \emph{frogs}, \emph{horses}, \emph{ships}, and \emph{trucks}. The training and test sets contain 50,000 and 10,000 images respectively. 
The SVHN dataset contains colored digit images 0 to 9 with resolution 32 × 32. 
The training and test sets contain 604,388 and 26,032 images respectively. 
For both CIFAR10 and SVHN, the training set is used to train and modularize models, and the test set is used to evaluate the trained models and the resulting modules.

\textbf{Baselines.} (i) Standard training. Standard training optimizes CNN models using mini-batch stochastic gradient descent with cross-entropy loss.
(ii) CNNSplitter~\cite{qi2022patching}. CNNSplitter is the state-of-the-art approach following the paradigm of modularizing after training. In particular, same as \projectNameCNN, CNNSplitter also produces modules by retaining relevant convolution kernels.

\textbf{Metrics.} (i) Classification accuracy (ACC), which is calculated as the percentage of correct predictions on the test set out of the total number of predictions made on the test set. (ii) Kernel retention rate (KRR)~\cite{qi2022patching}, which is calculated as the average number of kernels retained in modules divided by the total number of kernels in the model. (iii) Cohesion, which is the average cohesion of all modules (Eq. \ref{eq:cohesion}). (iv) Coupling, which is the average coupling across all pairs of modules (Eq. \ref{eq:coupling}). 

\textbf{Hyper-parameters.}
In standard training, VGG16 and ResNet18 are trained for 200 epochs using a mini-batch size of 128. We set the learning rate to 0.05 and Nesterov's momentum to 0.9. Additionally, we use common data augmentation~\cite{augmentation}, such as random cropping and flipping. As for SimCNN and ResCNN, pre-trained models published by the baseline~\cite{qi2022patching} are utilized directly. For \projectNameCNN, the hyperparameters involved in modular training include all the hyperparameters in standard training, and their settings are the same as those in standard training. In addition, the settings of weighting factors $\alpha$ and $\beta$ are shown in Table \ref{tab:alpha_beta}, and the threshold $\tau$ is set to 0.9. The effect of $\alpha$, $\beta$, and $\tau$ will be investigated in RQ4.

\begin{table}[t]
\setlength\tabcolsep{3pt}
\caption{The settings of weighting factors $\alpha$ and $\beta$.}
\label{tab:alpha_beta}
\vspace{-6pt}
\centering
\resizebox{\columnwidth}{!}{
\begin{tabular}{ccccccccc}
\toprule
               & \multicolumn{2}{c}{\textbf{VGG16}} & \multicolumn{2}{c}{\textbf{ResNet18}} & \multicolumn{2}{c}{\textbf{SimCNN}} & \multicolumn{2}{c}{\textbf{ResCNN}} \\ \cmidrule(lr){2-3} \cmidrule(lr){4-5} \cmidrule(lr){6-7} \cmidrule(lr){8-9}
               & \textbf{CIFAR10}  & \textbf{SVHN}  & \textbf{CIFAR10}    & \textbf{SVHN}   & \textbf{CIFAR10}   & \textbf{SVHN}  & \textbf{CIFAR10}   & \textbf{SVHN}  \\ \midrule \midrule
\textbf{$\alpha$} & 0.5               & 1.0              & 0.5                 & 1.3             & 0.5                & 0.5            & 1.0                  & 1.0              \\ \hline
\textbf{$\beta$}  & 1.5               & 2.0              & 1.5                 & 1.5             & 3.5                & 2.0              & 2.2                & 1.8          \\ \bottomrule
\end{tabular}
}

\end{table}

All the experiments are conducted on Ubuntu 20.04 server with 64 cores of 2.3GHz CPU, 128GB RAM, and NVIDIA Ampere A100 GPUs with 40 GB memory.

\subsection{Experimental Results}

\label{subsec:results}
\textbf{RQ1: How effective is \projectNameCNN in training and modularizing CNN models?}

\begin{table}[t]
\setlength\tabcolsep{2pt}
\caption{The comparison of standard training and the proposed \projectNameCNN.}
\label{tab:rq1_st_and_ours}
\vspace{-6pt}
\centering
\resizebox{\columnwidth}{!}{
\begin{tabular}{cccccccc}
\toprule
\multirow{2}{*}{\textbf{Model}} & \multirow{2}{*}{\textbf{Dataset}} & \multirow{2}{*}{\textbf{\#Kernels}} & \multirow{2}{*}{\textbf{\begin{tabular}[c]{@{}c@{}}Standard\\  Model ACC\end{tabular}}} & \multirow{2}{*}{\textbf{\begin{tabular}[c]{@{}c@{}}Modular\\ Model ACC\end{tabular}}} & \multicolumn{3}{c}{\textbf{Modules}}                      \\ \cmidrule(lr){6-8}
                                &                                   &                                      &                                                                                         &                                                                                       & \textbf{KRR} & \textbf{Cohesion} & \textbf{Coupling} \\ \midrule \midrule
\multirow{2}{*}{VGG16}          & CIFAR10                           & \multirow{2}{*}{4224}                & 92.29                                                                                   & 90.86                                                                                 & 17.28             & 0.9758            & 0.1751            \\
                                & SVHN                              &                                      & 95.84                                                                                   & 94.74                                                                                 & 14.15             & 0.9687            & 0.2246            \\ \midrule
\multirow{2}{*}{ResNet18}       & CIFAR10                           & \multirow{2}{*}{3904}                & 93.39                                                                                   & 91.59                                                                                 & 24.74             & 0.9437            & 0.2412            \\
                                & SVHN                              &                                      & 95.84                                                                                   & 95.95                                                                                 & 25.89             & 0.9663            & 0.3115            \\ \midrule
\multicolumn{2}{c}{\textbf{Average}}                                & \textbf{4064}                        & \textbf{94.34}                                                                          & \textbf{93.29}                                                                        & \textbf{20.52}    & \textbf{0.9636}   & \textbf{0.2381}   \\ \bottomrule
\end{tabular}
}
\end{table}

To evaluate the effectiveness of \projectNameCNN in training and modularizing CNN models, we evaluate the classification accuracy of modular models and measure the kernel retention rate (KRR), cohesion, and coupling of trained modules.
Table \ref{tab:rq1_st_and_ours} presents the results of standard training, modular training, and modularizing on four models. Column ``\#Kernels'' shows the number of convolution kernels for VGG16 and ResNet18, and the fourth and fifth columns show the accuracy of standard and modular models on testing set, respectively. 
On average, the standard and modular models achieved accuracy of 94.34\% and 93.29\%, respectively. 
The average loss of accuracy is 1.05 percentage points, demonstrating that modular training does not cause much loss of performance in terms of accuracy.
The last three columns display KRR, cohesion and coupling degree of obtained modules by modularizing with a threshold of 0.9, respectively.
On average, the modules of all four models retain 20.52\% of convolution kernels.
The averaged cohesion degree is 0.9636, indicating that a module uses essentially the same convolution kernels to predict samples belonging to the corresponding class. 
The averaged coupling between modules is 0.2381, indicating that the convolution kernels of different modules do not overlap much.

\begin{figure}
    \centering
    \includegraphics[width=\columnwidth]{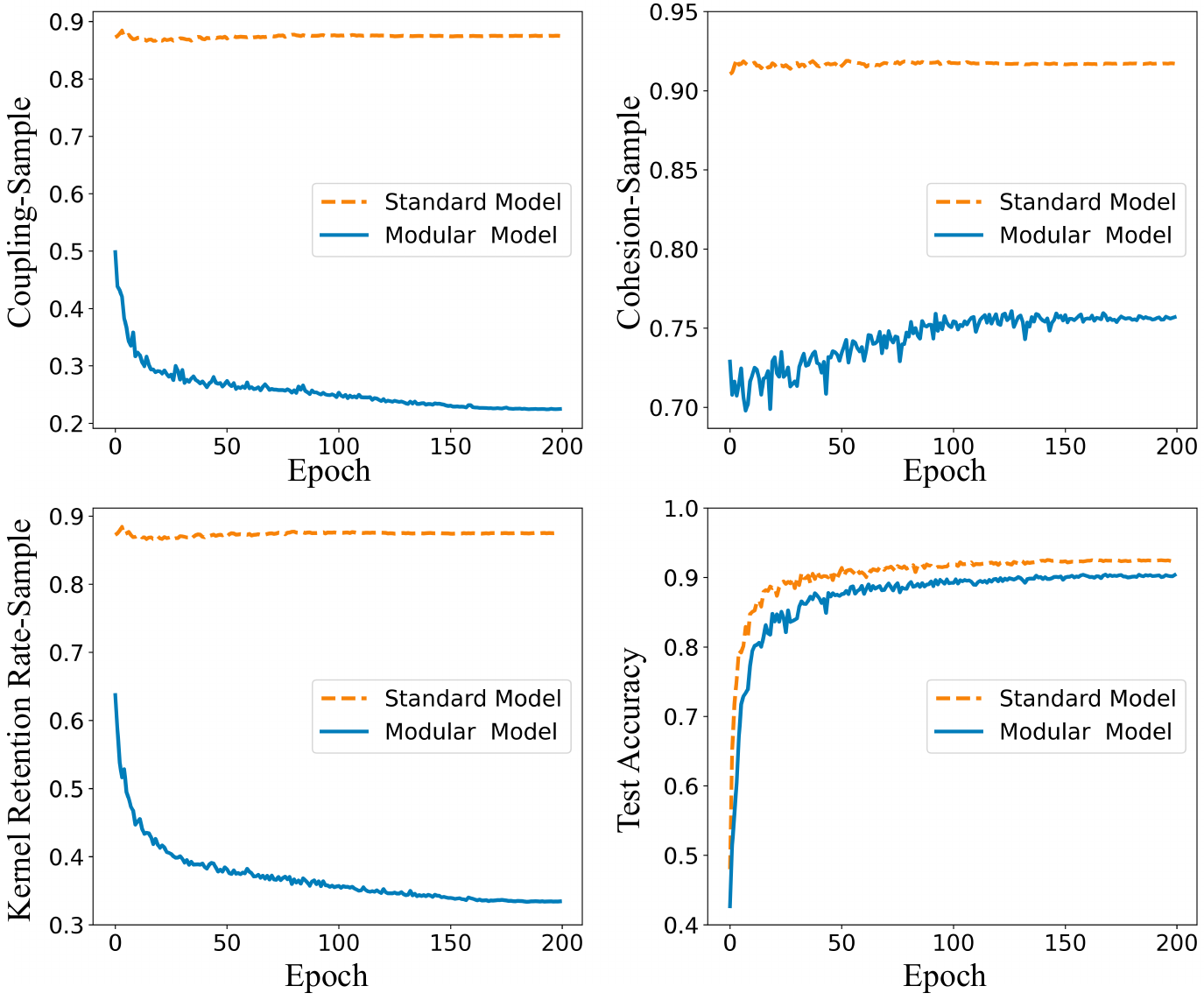}
    \caption{The convergence process of Modular Training on VGG16-CIFAR10.}
    \label{fig:rq1_convergence}
    \vspace{-10pt}
\end{figure}

More specifically, Figure \ref{fig:rq1_convergence} depicts the convergence process of training for the VGG16 model on the CIFAR10 dataset, where the dashed orange lines represent the standard training and solid blue lines represent modular training.
During modular training, the coupling and cohesion of the modular model in each epoch are calculated as the average values across all iterations within that epoch. The coupling and cohesion in each iteration are calculated using Eq. \ref{eq:coupling} and Eq. \ref{eq:cohesion}.
Note that, the calculation of coupling and cohesion during modular training does not involve filtering, as the batch of samples (consisting of 128 samples) in each iteration may not accurately estimate the occurrence frequency of a kernel in the entire training dataset.
Regarding the coupling and cohesion of the standard model during standard training, relevant kernels for a sample are identified based on whether their outputs contain non-zero elements. If a kernel's output contains non-zero elements, it is considered relevant because its output may affect the classification of the sample. Otherwise, the kernel is considered irrelevant. 
Overall, the process of identifying relevant kernels during standard training is similar to that during modular training. In both cases, the determination is based on whether the output of a kernel will impact the classification of the sample.

The top left sub-figure plots the convergence trend of coupling degree.
Regarding the modular model, the coupling decreases rapidly in the first 50 epochs and gradually converges. In contrast, the coupling degree of the standard model is consistently high. 
This indicates that the proposed loss function can effectively reduce the coupling degree between kernels that are relevant to different classes during the training process.
The top right sub-figure shows the convergence process of cohesion degree. 
The cohesion degrees of both modular and standard models are maintained at a relatively high level.
However, this does not imply that the designed cohesion loss function is unimportant.
In our experiments, when eliminating the cohesion loss, the cohesion degree of the modular model would quickly drop as the coupling degree decreases.
The reduction of the cohesion degree would result in the increase of module size, thus increasing the reuse overhead of the modules.
Compared to the standard training that maintains both high cohesion and coupling degree, \projectNameCNN is able to maintain a high cohesion degree while reducing the coupling degree.
When predicting samples belonging to different classes, the standard model always uses most of the convolution kernels, while the modular model uses only a corresponding portion of the convolution kernels. 

The bottom left sub-figure gives the averaged rate of convolution kernels used for predicting samples. 
The standard model always uses the majority of convolution kernels to make predictions. In contrast, the modular model tends to use as few convolution kernels as possible, thus reducing the coupling degree. 
The fewer convolution kernels used for predicting a sample, the less likely that the convolution kernels for different prediction tasks overlap. 
Similar to the convergence trend of coupling degree, the KRR decreases rapidly in the first 50 epochs and gradually converges.

Despite the differences in cohesion and coupling degrees, the modular model achieves competitive classification accuracy with the standard model.
The convergence trends of the accuracy of the modular and standard models on the test dataset are similar, with final classification accuracy of 90.86\% and 92.29\%, respectively.
Modular training results in a small accuracy loss of 1.43 percentage points.

\begin{table}[t]
\setlength\tabcolsep{2pt}
\caption{The comparison of CNNSplitter and  \projectNameCNN. ``S.M.'', ``Coup.'', and ``Cohe.'' indicate ``Standard Model ACC'', ``Coupling'', and ``Cohesion'', respectively.}
\label{tab:rq1_cnnsplitter_and_ours}
\vspace{-6pt}
\centering
\resizebox{\columnwidth}{!}{
\begin{tabular}{ccccccccccc}
\toprule
\multirow{2}{*}{\textbf{Model}}  & \multirow{2}{*}{\textbf{Dataset}} & \multirow{2}{*}{\textbf{\begin{tabular}[c]{@{}c@{}}S.M. \\ ACC\end{tabular}}} & \multicolumn{4}{c}{\textbf{CNNSplitter}}                                   & \multicolumn{4}{c}{\textbf{\projectNameCNN}}                                          \\ \cmidrule(lr){4-7} \cmidrule(lr){8-11} 
                                 &                                   &                                                                                         & \textbf{ACC}   & \textbf{Coup.} & \textbf{Cohe.} & \textbf{KRR} & \textbf{ACC}   & \textbf{Coup.} & \textbf{Cohe.} & \textbf{KRR} \\ \midrule \midrule
\multirow{2}{*}{SimCNN} & CIFAR10                  & 89.77                                                                                   & 86.07          & 0.5277            & 0.9326            & 61.96             & 88.84          & 0.1372            & 0.8682            & 11.58             \\ \cmidrule(lr){2-11} 
                                 & SVHN                     & 95.41                                                                                   & 93.85          & 0.6161            & 0.9619            & 52.79             & 93.56          & 0.1434            & 0.9580            & 11.85             \\ \midrule
\multirow{2}{*}{ResCNN} & CIFAR10                  & 90.41                                                                                   & 85.64          & 0.5648            & 0.8462            & 58.26             & 89.82          & 0.2781            & 0.9601            & 21.52             \\ \cmidrule(lr){2-11} 
                                 & SVHN                     & 95.06                                                                                   & 93.52          & 0.6046            & 0.8828            & 54.03             & 93.88          & 0.3306            & 0.9731            & 13.37             \\ \midrule
\multicolumn{2}{c}{\textbf{Average}}                                 & \textbf{92.66}                                                                          & \textbf{89.77} & \textbf{0.5783}   & \textbf{0.9059}   & \textbf{56.76}    & \textbf{91.53} & \textbf{0.2223}   & \textbf{0.9399}   & \textbf{14.58}    \\ \bottomrule
\end{tabular}
}
\end{table}

Moreover, we also compare \projectNameCNN with the state-of-the-art modularization approach CNNSplitter~\cite{qi2022patching}. 
Specifically, we directly compared these two approaches on the models and modules published by CNNSplitter.
Table~\ref{tab:rq1_cnnsplitter_and_ours} shows the performance on four metrics, including accuracy, kernel retention rate, cohesion, and coupling.
Note that, CNNSplitter's ACC represents the test accuracy of the ``composed'' model constructed by combining all the decomposed modules. 
The composed model is similar to the modular model.
On average, the accuracy of CNNSplitter's composed model and \projectNameCNN's modular model are 89.77\% and 91.53\%, respectively, with the latter achieving an improvement of 1.76 percentage points. Compared with the accuracy of the standard model of 92.66\%, the accuracy losses caused by CNNSplitter and \projectNameCNN are 2.89 percentage points and 1.13 percentage points, respectively, indicating that \projectNameCNN causes much less loss of accuracy. 
To compare cohesion and coupling, we calculate the cohesion degree of CNNSplitter's modules in the same way as standard model. 
Since a module generated by CNNSplitter retains only a part of the convolution kernels of the standard model,
the calculation of the coupling degree of CNNSplitter's modules is the same as that of \projectNameCNN, which is obtained by calculating the Jaccard similarity between modules. 
Overall, compared with CNNSplitter, \projectNameCNN achieves better results in both coupling and cohesion metrics, with an average improvement of 0.3560 and 0.0340, respectively. Regarding convolution kernel retention rate, %
\projectNameCNN is significantly better than CNNSplitter (56.76\% vs 14.58\%), with a reduction of 74.31\% in KRR.

Similar to \projectNameCNN, CNNSplitter uses Jaccard similarity to increase the differences between modules and ensure modules retain only relevant convolution kernels. However, CNNSplitter is an approach that modularizes after training. Standard training methods do not consider the differences (i.e., coupling) between modules, which results in small differences between modules. Therefore, modularizing-after-training approaches are inherently limited in terms of coupling and KRR. In contrast, \projectNameCNN is a modularizing-while-training approach, which considers reducing the coupling between modules during the modular training process, leading to better results in terms of coupling and KRR.

\begin{tcolorbox}[left=2pt,right=2pt,top=2pt,bottom=2pt]
The coupling, cohesion, and KRR of the modules generated by \projectNameCNN are 0.2223, 0.9399, and 14.58\%, respectively, which are decreased by 0.3560, increased by 0.0340, and reduced by 74.31\%, respectively, compared to the state-of-the-art.
\end{tcolorbox}

\textbf{RQ2: How efficient is \projectNameCNN in training and modularizing CNN models?}

\begin{table}[t]
\setlength\tabcolsep{2.5pt}
\caption{The comparison of \projectNameCNN and modularizing-after-training in time costs of training and modularizing. ``Mo.'' indicates ``Modularizing''.}
\label{tab:rq3_timecost}
\vspace{-6pt}
\centering
\resizebox{\columnwidth}{!}{
\begin{tabular}{ccrrrrrr}
\toprule
\multirow{3}{*}{\textbf{Model}}  & \multirow{3}{*}{\textbf{Dataset}} & \multicolumn{3}{c}{\textbf{MwT}}                                                                             & \multicolumn{3}{c}{\textbf{Modularizing-after-Training}}                                                     \\ \cmidrule(lr){3-5} \cmidrule(lr){6-8} 
                                 &                                   & \textbf{\begin{tabular}[c]{@{}c@{}}Modular\\ Training\end{tabular}} & \textbf{Mo.} & \textbf{Total} & \textbf{\begin{tabular}[c]{@{}c@{}}Standard\\ Training\end{tabular}} & \textbf{CNNSplitter} & \textbf{Total} \\ \midrule \midrule
\multirow{2}{*}{SimCNN} & CIFAR10                  & 24s/e x 200e                                                        & 14s                   & 80m            & 12s/e x 200e                                                         & 83s/e x 123e         & 210m           \\
                                 & SVHN                     & 36s/e x 200e                                                        & 21s                   & 120m           & 17s/e x 200e                                                         & 95s/e x   79e        & 182m           \\ \midrule
\multirow{2}{*}{ResCNN} & CIFAR10                  & 28s/e x 200e                                                        & 15s                   & 94m            & 14s/e x 200e                                                         & 80s/e x 185e         & 293m           \\
                                 & SVHN                     & 41s/e x 200e                                                        & 21s                   & 137m           & 20s/e x 200e                                                         & 93s/e x 107e         & 233m           \\ \midrule
\multicolumn{2}{c}{\textbf{Average}}                                 & \textbf{107m}                                                       & \textbf{18s}          & \textbf{108m}  & \textbf{53m}                                                         & \textbf{177m}        & \textbf{229m}  \\ \bottomrule
\end{tabular}
}
\end{table}

One advantage of \projectNameCNN over Modularizing-after-training is to reduce the runtime overhead. 
In this experiment, we take CNNSplitter as the example of modularizing-after-training.
The runtime overhead of \projectNameCNN includes modular training time and modularizing time, and the runtime overhead of CNNSplitter includes standard training time and modularizing time.
Table \ref{tab:rq3_timecost} shows the runtime overhead for \projectNameCNN and modularizing-after-training on SimCNN-CIFAR10, SimCNN-SVHN, ResCNN-CIFAR10, and ResCNN-SVHN. 
For instance, in the case of SimCNN-CIFAR10, the modular training time of \projectNameCNN amounts to $24s/e \times 200e$, indicating that the model is trained for 200 epochs, with each epoch taking 24 seconds. 
The modularizing time of \projectNameCNN is mainly attributed to the generation of masks through forward propagation, which takes 14 seconds. 
Consequently, the runtime overhead of \projectNameCNN amounts to 80 minutes.
Regarding CNNSplitter, the standard training time amounts to $12s/e \times 200e$.
The modularizing time of CNNSplitter amounts to $83s/e \times 123e$, i.e., the modularization iterates 123 epochs, with each epoch taking 83 seconds. 
As a result, the runtime overhead of modularizing-after-training amounts to 210 minutes. On average, \projectNameCNN's runtime overhead is 108 minutes, while modularizing-after-training's runtime overhead is 229 minutes. 

Additionally, we found that the average modular training time of \projectNameCNN (i.e., 107 minutes) is almost twice the standard training time (i.e., 53 minutes). This is mainly due to the introduction of additional parameters (i.e., mask generators) and loss functions (i.e., cohesion loss and coupling loss). %
Moreover, in our experiments, we noticed that the GPU utilization for modular training is less than 70\%, while the GPU utilization for standard training is nearly 100\%. 
This suggests that optimizing the implementation of \projectNameCNN could further enhance the GPU utilization, thereby reducing the time overhead of modular training. 
We leave this as future work.

\begin{tcolorbox}[left=2pt,right=2pt,top=2pt,bottom=2pt]
On average, the time overhead of \projectName is 108 minutes, only half of the time for modularizing-after-training.
\end{tcolorbox}

\textbf{RQ3: How effective is \projectNameCNN in reusing CNN modules?}

In this RQ, we investigate the effectiveness of \projectNameCNN in on-demand model reuse, which is one of the key benefits of model modularization.
Specifically, we reuse both the module and the standard model on sub-tasks derived from the original classification task that a standard model solves. %
There are two 10-class classification tasks corresponding to the CIFAR10 and SVHN datasets. Each task can be divided into sub-tasks with a number of categories ranging from 2 to 10. 
An $M$-class sub-task consists of $M$ categories from the 10-class classification task, resulting in a total of $C_{10}^{M}$ $M$-class sub-tasks. 
For instance, a 5-class sub-task has $252$ ($C_{10}^{5}$) possibilities in total.
We randomly select 10 sub-tasks for each 2-class to 9-class classification sub-task. 
For each $M$-class sub-task, we reuse the standard model and corresponding modules (see Sec. \ref{subsec:reuse}) that can classify the $M$ categories, then compare the number of convolution kernels, computational cost, and accuracy of model and modules.

\begin{table}[t]
\caption{The convolution kernel retention rate of \projectNameCNN in reusing CNN modules. All results in \%.}
\label{tab:rq2_standard_and_ours_krr}
\vspace{-6pt}
\centering
\resizebox{0.9\columnwidth}{!}{
\begin{tabular}{cccccc}
\toprule
\multirow{2}{*}{\textbf{Target Task}} & \multicolumn{2}{c}{\textbf{VGG16}} & \multicolumn{2}{c}{\textbf{ResNet18}} & \multirow{2}{*}{\textbf{Average}} \\ \cmidrule(lr){2-3} \cmidrule(lr){4-5}
                                      & \textbf{CIFAR10}  & \textbf{SVHN}  & \textbf{CIFAR10}    & \textbf{SVHN}   &                                   \\ \midrule \midrule
2-class                      & 30.34             & 27.38          & 35.94               & 42.44           & 34.03                             \\
3-class                      & 43.39             & 38.65          & 50.51               & 68.18           & 50.18                             \\
4-class                      & 52.93             & 47.00          & 65.12               & 70.38           & 58.86                             \\
5-class                      & 58.29             & 53.15          & 71.41               & 72.28           & 63.78                             \\
6-class                      & 63.75             & 57.31          & 74.58               & 74.01           & 67.41                             \\
7-class                      & 72.57             & 61.27          & 79.79               & 78.36           & 73.00                             \\
8-class                      & 79.88             & 63.65          & 82.35               & 79.56           & 76.36                             \\
9-class                      & 88.34             & 66.25          & 86.87               & 80.60           & 80.52                             \\ 
10-class                     & 94.11             & 68.61          & 88.75               & 81.52           & 83.25                             \\ \bottomrule
\end{tabular}
}
\end{table}

Table \ref{tab:rq2_standard_and_ours_krr} presents the KRR of the modules on different classification sub-tasks.
For instance, as shown in the third row, for a 2-class classification sub-task of CIFAR10, the corresponding modules of VGG16-CIFAR10 only retain 30.34\% of the model's convolution kernels. 
On average, for 2-class classification sub-tasks, the corresponding modules retain only 34.03\% of the kernels. 
As the number of categories of sub-tasks increases, the number of kernels retained by the corresponding modules also increases.
However, we found that even when the number of categories increases to the maximum (i.e., the sub-task is the original task), the corresponding modules contain fewer kernels than the model, with an average retention rate of 83.25\%. The reason is that the \textit{modularizing} stage filters out unimportant convolution kernels (see Sec. \ref{subsec:modularizing}).

\begin{table}[t]
\setlength\tabcolsep{3pt}
\caption{The computational costs of \projectNameCNN in reusing CNN modules.}
\label{tab:rq2_standard_and_ours_flops}
\vspace{-6pt}
\centering
\resizebox{\columnwidth}{!}{
\begin{tabular}{ccrrrrr}
\toprule
\multirow{2}{*}{\textbf{Target Task}} & \multirow{2}{*}{\textbf{Metric}} & \multicolumn{2}{c}{\textbf{VGG16}} & \multicolumn{2}{c}{\textbf{ResNet18}} & \multirow{2}{*}{\textbf{Average}} \\ \cmidrule(lr){3-4} \cmidrule(lr){5-6}
                                      &                                  & \textbf{CIFAR10}  & \textbf{SVHN}  & \textbf{CIFAR10}    & \textbf{SVHN}   &                                   \\ \midrule \midrule
\multirow{2}{*}{2-class}     & FLOPs (M)               & 70.97             & 31.85          & 129.76              & 136.15          & 92.18                             \\
                                      & Reduction (\%)          & 77.48             & 89.89          & 76.77               & 75.62           & 79.94                             \\ \hline
\multirow{2}{*}{3-class}     & FLOPs (M)               & 121.21            & 52.93          & 189.50              & 329.49          & 173.28                            \\
                                      & Reduction (\%)          & 61.54             & 83.20          & 66.07               & 41.00           & 62.95                             \\ \hline
\multirow{2}{*}{4-class}     & FLOPs (M)               & 166.35            & 73.31          & 336.49              & 336.41          & 228.14                            \\
                                      & Reduction (\%)          & 47.21             & 76.73          & 39.75               & 39.77           & 50.87                             \\ \hline
\multirow{2}{*}{5-class}     & FLOPs (M)               & 179.93            & 85.82          & 360.62              & 344.77          & 242.79                            \\
                                      & Reduction (\%)          & 42.90             & 72.77          & 35.43               & 38.27           & 47.34                             \\ \hline
\multirow{2}{*}{6-class}     & FLOPs (M)               & 198.71            & 91.03          & 376.75              & 352.28          & 254.69                            \\
                                      & Reduction (\%)          & 36.94             & 71.11          & 32.54               & 36.93           & 44.38                             \\ \hline
\multirow{2}{*}{7-class}     & FLOPs (M)               & 230.41            & 99.47          & 398.82              & 372.42          & 275.28                            \\
                                      & Reduction (\%)          & 26.88             & 68.43          & 28.59               & 33.32           & 39.31                             \\ \hline
\multirow{2}{*}{8-class}     & FLOPs (M)               & 249.33            & 103.30         & 407.57              & 377.54          & 284.44                            \\
                                      & Reduction (\%)          & 20.87             & 67.22          & 27.03               & 32.40           & 36.88                             \\ \hline
\multirow{2}{*}{9-class}     & FLOPs (M)               & 279.40            & 107.36         & 432.79              & 382.93          & 300.62                            \\
                                      & Reduction (\%)          & 11.33             & 65.93          & 22.51               & 31.44           & 32.80                             \\ \hline
\multirow{2}{*}{10-class}    & FLOPs (M)               & 293.88            & 111.10         & 442.67              & 388.09          & 308.94                            \\
                                      & Reduction (\%)          & 6.74              & 64.74          & 20.74               & 30.51           & 30.68                             \\ \bottomrule
\end{tabular}
}
\end{table}

Lower KRR means that on-demand model reuse incurs a lower computational cost.
To compare the computational cost directly, we follow the baseline~\cite{qi2022patching} and use the open-source tool fvcore~\cite{fvcore} to calculate the number of floating point operations (FLOPs) required by the model and module to make predictions. %
The VGG16 and ResNet18 models require 315.11 million and 558.50 million FLOPs, respectively. 
Table \ref{tab:rq2_standard_and_ours_flops} displays the number of FLOPs (million) required by modules, %
as well as the percentage reduction (\%) compared to reusing the model.
For instance, the VGG16-CIFAR10 modules for 2-class sub-tasks require 70.97 million FLOPs, resulting in a cost reduction of 77.48\% ($(1 - 70.97/315.11) \times 100$) compared to the original VGG16 model.
Reusing the modules of ResNet18-CIFAR10 for 2-class sub-tasks requires 129.76 million FLOPs, resulting in a 76.77\% cost reduction compared to reusing the model. 
As shown in the last column, on average, reusing modules reduces the computational cost by 79.94\% for 2-class classification sub-tasks.
Overall, the experimental results demonstrate that on-demand model reuse significantly reduces the reuse cost. 
In particular, the average computational cost can be reduced by more than 50\% when the number of categories in a sub-task is less than half of the original task.

\begin{table}[t]
\caption{The test accuracy results of \projectNameCNN in reusing CNN modules. All results in \%.}
\label{tab:rq2_standard_and_ours_acc}
\vspace{-6pt}
\centering
\resizebox{\columnwidth}{!}{
\begin{tabular}{ccrrrrr}
\toprule
\multirow{2}{*}{\textbf{Target Task}} & \multirow{2}{*}{\textbf{Metric}} & \multicolumn{2}{c}{\textbf{VGG16}} & \multicolumn{2}{c}{\textbf{ResNet18}} & \multirow{2}{*}{\textbf{Average}} \\ \cmidrule(lr){3-4} \cmidrule(lr){5-6}
                                      &                                  & \textbf{CIFAR10}  & \textbf{SVHN}  & \textbf{CIFAR10}    & \textbf{SVHN}   &                                   \\ \midrule \midrule
\multirow{3}{*}{2-class}     & M-ACC                   & 99.35             & 98.89          & 99.40               & 99.34           & 99.25                             \\
                                      & m-ACC                   & 99.20             & 99.17          & 99.30               & 99.33           & 99.25                             \\ \cline{2-7} 
                                      & \textbf{loss}                    & 0.15              & -0.28          & 0.10                & 0.01            & 0.00                              \\ \hline
\multirow{3}{*}{3-class}     & M-ACC                   & 98.23             & 98.48          & 98.23               & 98.91           & 98.46                             \\
                                      & m-ACC                   & 97.40             & 98.58          & 97.23               & 98.82           & 98.01                             \\ \cline{2-7} 
                                      & \textbf{loss}                    & 0.83              & -0.10          & 1.00                & 0.09            & 0.45                              \\ \hline
\multirow{3}{*}{4-class}     & M-ACC                   & 96.70             & 97.35          & 96.63               & 97.96           & 97.16                             \\
                                      & m-ACC                   & 94.98             & 97.28          & 95.50               & 97.64           & 96.35                             \\ \cline{2-7} 
                                      & \textbf{loss}                    & 1.72              & 0.07           & 1.13                & 0.32            & 0.81                              \\ \hline
\multirow{3}{*}{5-class}     & M-ACC                   & 95.22             & 97.01          & 95.66               & 97.47           & 96.34                             \\
                                      & m-ACC                   & 93.82             & 97.01          & 94.34               & 97.55           & 95.68                             \\ \cline{2-7} 
                                      & \textbf{loss}                    & 1.40              & 0.00           & 1.32                & -0.08           & 0.66                              \\ \hline
\multirow{3}{*}{6-class}     & M-ACC                   & 92.53             & 96.73          & 93.77               & 96.98           & 95.00                             \\
                                      & m-ACC                   & 91.05             & 96.44          & 91.27               & 96.92           & 93.92                             \\ \cline{2-7} 
                                      & \textbf{loss}                    & 1.48              & 0.29           & 2.50                & 0.06            & 1.08                              \\ \hline
\multirow{3}{*}{7-class}     & M-ACC                   & 92.43             & 96.44          & 93.60               & 96.74           & 94.80                             \\
                                      & m-ACC                   & 90.90             & 95.78          & 91.86               & 96.45           & 93.75                             \\ \cline{2-7} 
                                      & \textbf{loss}                    & 1.53              & 0.66           & 1.74                & 0.29            & 1.06                              \\ \hline
\multirow{3}{*}{8-class}     & M-ACC                   & 92.28             & 96.18          & 93.48               & 96.32           & 94.57                             \\
                                      & m-ACC                   & 91.03             & 95.52          & 91.71               & 96.25           & 93.63                             \\ \cline{2-7} 
                                      & \textbf{loss}                    & 1.25              & 0.66           & 1.77                & 0.07            & 0.94                              \\ \hline
\multirow{3}{*}{9-class}     & M-ACC                   & 92.36             & 96.04          & 93.52               & 96.09           & 94.50                             \\
                                      & m-ACC                   & 90.89             & 95.20          & 91.48               & 96.00           & 93.39                             \\ \cline{2-7} 
                                      & \textbf{loss}                    & 1.47              & 0.84           & 2.04                & 0.09            & 1.11                              \\ \hline
\multirow{3}{*}{10-class}    & M-ACC                   & 92.29             & 95.84          & 93.39               & 95.84           & 94.34                             \\
                                      & m-ACC                   & 90.86             & 94.74          & 91.59               & 95.95           & 93.29                             \\ \cline{2-7} 
                                      & \textbf{loss}                    & 1.43              & 1.10           & 1.80                & -0.11           & 1.06                              \\ \bottomrule
\end{tabular}
}
\end{table}

In practice, on-demand model reuse must balance the need to reduce reuse overhead and the need to maintain classification accuracy.
Table \ref{tab:rq2_standard_and_ours_acc} presents the accuracy of both the standard model and modules on the sub-tasks (i.e., \emph{M-ACC} and \emph{m-ACC}).
Meanwhile, row \textit{loss} presents the accuracy loss caused by module reuse, which is the difference in accuracy between the module and the model.
Evaluation results show that module reuse causes more accuracy loss in sub-tasks with over half of the original task's categories, e.g., the modules for 9-class sub-tasks result in an accuracy loss of approximately 1 percentage point. 
In contrast, sub-tasks with less than half categories have a smaller accuracy loss (<1 percentage point). 
The reason could be that as the category number increases, the classification task becomes more challenging, and it requires more classifier's parameters.
Furthermore, the table indicates that the accuracy loss is higher for CIFAR10 sub-tasks compared to SVHN sub-tasks, possibly due to the greater complexity of the CIFAR10 task, which requires a classifier to be equipped with more parameters.
Overall, on-demand reuse of modules causes a negligible average accuracy loss of only 0.8 percentage points across all cases, demonstrating that \projectNameCNN can strike a balance between classification accuracy and computational cost.

\begin{table}[t]
\setlength\tabcolsep{3pt}
\caption{The comparison of \projectNameCNN and CNNSplitter in reusing CNN modules in terms of KRR. All results in \%.}
\label{tab:rq2_cnnsplitter_and_ours_krr}
\vspace{-6pt}
\centering
\resizebox{\columnwidth}{!}{
\begin{tabular}{ccccccc}
\toprule
\multirow{2}{*}{\textbf{Target Task}} & \multirow{2}{*}{\textbf{Approach}} & \multicolumn{2}{c}{\textbf{SimCNN}} & \multicolumn{2}{c}{\textbf{ResCNN}} & \multirow{2}{*}{\textbf{Average}} \\ \cmidrule(lr){3-4} \cmidrule(lr){5-6}
                                      &                                    & \textbf{CIFAR10}   & \textbf{SVHN}  & \textbf{CIFAR10}   & \textbf{SVHN}  &                                   \\ \midrule \midrule
\multirow{2}{*}{2-class}     & CNNSplitter               & 84.61              & 76.04          & 79.57              & 78.26          & 79.62                             \\
                                      & \projectName                      & 20.34              & 34.13          & 28.69              & 19.98          & 25.79                             \\ \hline
\multirow{2}{*}{3-class}     & CNNSplitter               & 93.09              & 86.88          & 89.79              & 89.86          & 89.91                             \\
                                      & \projectName                      & 30.77              & 43.10          & 39.28              & 42.33          & 38.87                             \\ \hline
\multirow{2}{*}{4-class}     & CNNSplitter               & 97.09              & 93.42          & 95.22              & 94.73          & 95.12                             \\
                                      & \projectName                      & 33.80              & 51.49          & 53.46              & 45.64          & 46.10                             \\ \hline
\multirow{2}{*}{5-class}     & CNNSplitter               & 98.63              & 96.64          & 97.69              & 97.50          & 97.62                             \\
                                      & \projectName                      & 37.67              & 56.97          & 56.30              & 47.18          & 49.53                             \\ \hline
\multirow{2}{*}{6-class}     & CNNSplitter               & 99.34              & 98.37          & 98.79              & 98.72          & 98.81                             \\
                                      & \projectName                      & 41.25              & 64.69          & 58.62              & 49.34          & 53.48                             \\ \hline
\multirow{2}{*}{7-class}     & CNNSplitter               & 99.74              & 99.05          & 99.32              & 99.51          & 99.41                             \\
                                      & \projectName                      & 52.00              & 68.75          & 66.45              & 51.04          & 59.56                             \\ \hline
\multirow{2}{*}{8-class}     & CNNSplitter               & 99.88              & 99.43          & 99.77              & 99.81          & 99.72                             \\
                                      & \projectName                      & 57.14              & 72.55          & 68.37              & 52.37          & 62.61                             \\ \hline
\multirow{2}{*}{9-class}     & CNNSplitter               & 99.95              & 99.69          & 99.88              & 99.86          & 99.85                             \\
                                      & \projectName                      & 67.52              & 75.94          & 72.44              & 53.72          & 67.41                             \\ \hline
\multirow{2}{*}{10-class}    & CNNSplitter               & 99.98              & 99.76          & 99.91              & 99.88          & 99.88                             \\
                                      & \projectName                      & 72.40              & 78.82          & 74.22              & 54.71          & 70.04                             \\ \bottomrule
\end{tabular}
}
\end{table}

We further compare \projectNameCNN with CNNSplitter in terms of on-demand model reuse.
Similar to \projectNameCNN, which uses masks to represent modules, CNNSplitter uses a binary bit string to represent a module, where each bit represents whether the corresponding convolution kernel is retained or not.
By retaining kernels in the model based on bit vectors, the corresponding module can be constructed.
Table~\ref{tab:rq2_cnnsplitter_and_ours_krr} compares the results of KRR for CNNSplitter-based and \projectNameCNN-based module reuse.
The results show that \projectNameCNN achieves lower KRR than CNNSplitter on all sub-tasks.
For example, in the 2-class classification sub-task of CIFAR10, the KRR of the SimCNN-CIFAR10 modules constructed by \projectNameCNN and CNNSplitter are 20.34\% and 84.61\%, respectively.
The former contains significantly fewer kernels than the latter, resulting in a reduction of 75.96\%. 
On average, the modules constructed by \projectNameCNN retain 29.88\% (the 10-class sub-task) to 75.96\% (the 2-class sub-task) kernels less than the modules constructed by CNNSplitter. 
Moreover, it is worth mentioning that even for 2-class classification sub-tasks, the modules constructed by CNNSplitter retain most of the convolution kernels, with an average of 79.62\%, while the modules constructed by \projectNameCNN only retain 25.79\% of the convolution kernels. 
This is mainly because \projectNameCNN is a modularizing-while-training approach, while CNNSplitter relies on modularizing-after-training paradigm.
As a result, when reusing modules for $M$-class classification sub-tasks, the modules constructed by \projectNameCNN retain much fewer convolution kernels.

\begin{tcolorbox}[left=2pt,right=2pt,top=2pt,bottom=2pt]
Based on \projectNameCNN, reusing modules on-demand can significantly reduce the computational cost of reuse by up to 79.94\% with a negligible average accuracy loss of only 0.8 percentage points. 
\end{tcolorbox}

\textbf{RQ4: How do the major hyper-parameters influence the performance of \projectNameCNN?}

\begin{figure}[t]
\centering

\begin{minipage}{0.49\linewidth}
    \centering
    \includegraphics[width=\linewidth]{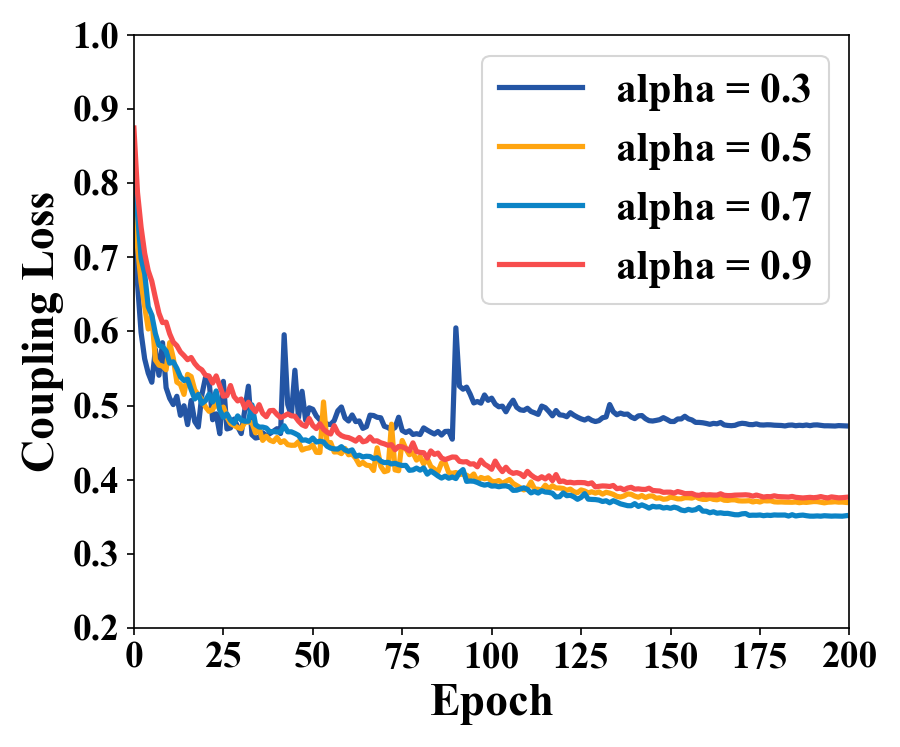}
\end{minipage}
\begin{minipage}{0.49\linewidth}
    \centering
    \includegraphics[width=\linewidth]{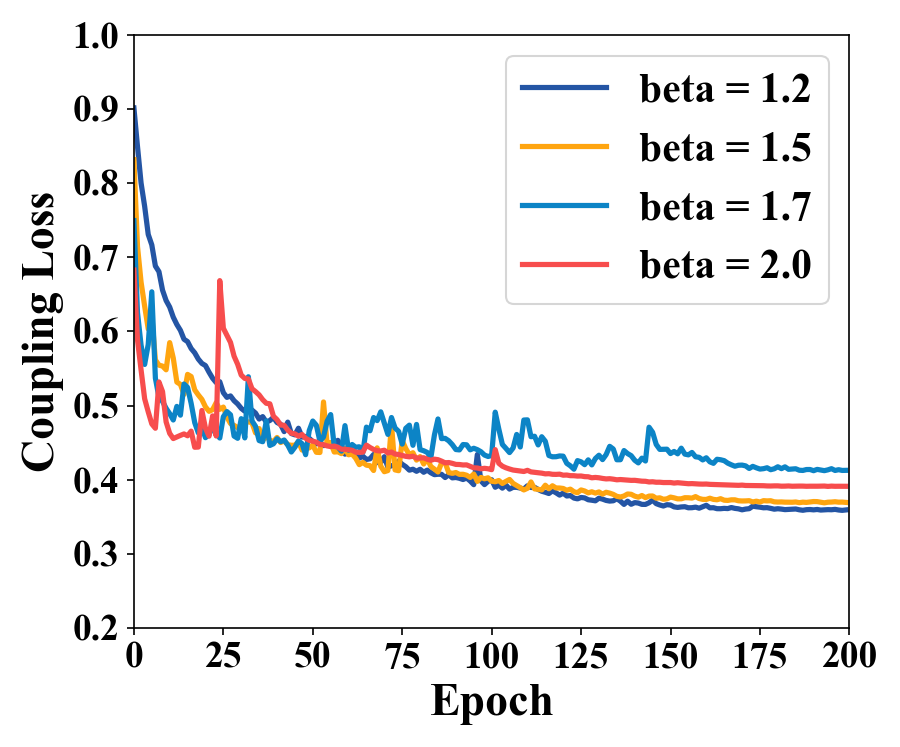}
\end{minipage}

\begin{minipage}{0.49\linewidth}
    \centering
    \includegraphics[width=\linewidth]{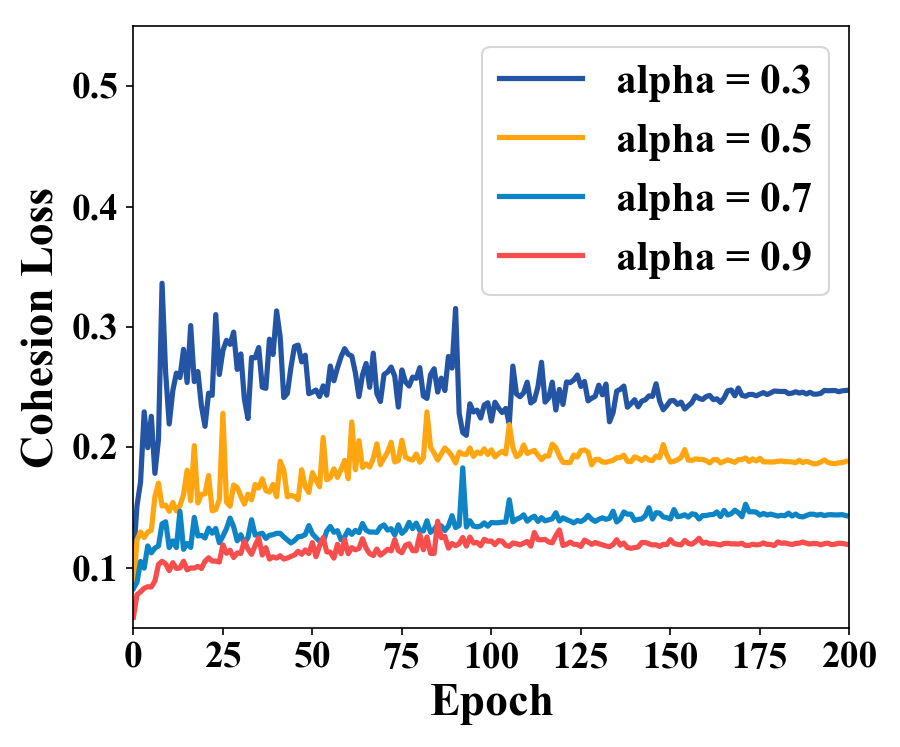}
\end{minipage}
\begin{minipage}{0.49\linewidth}
    \centering
    \includegraphics[width=\linewidth]{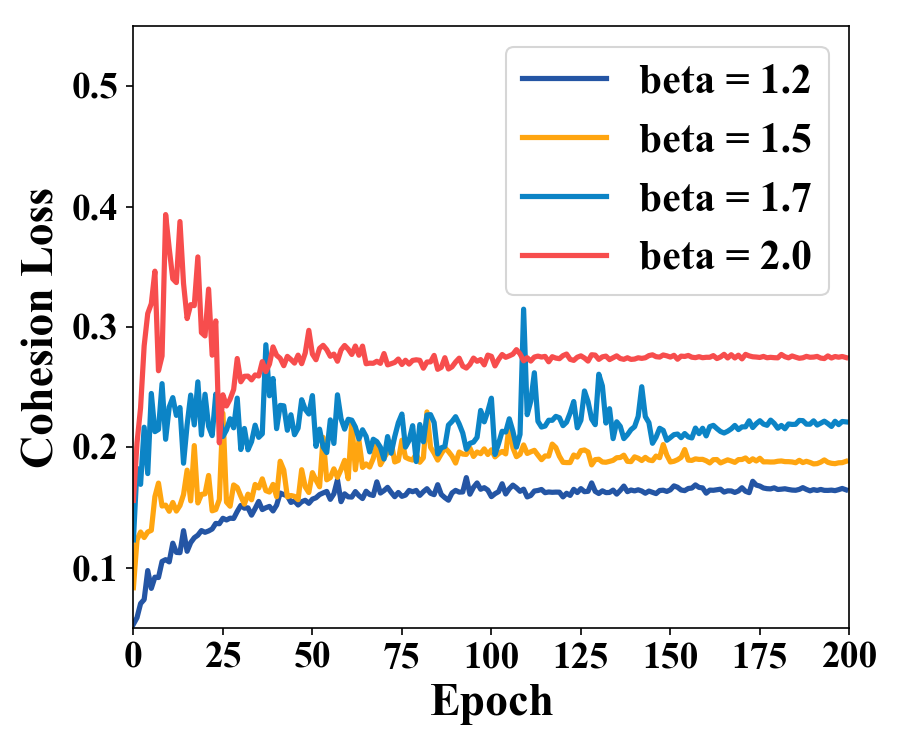}
\end{minipage}

\begin{minipage}{0.49\linewidth}
    \centering
    \includegraphics[width=\linewidth]{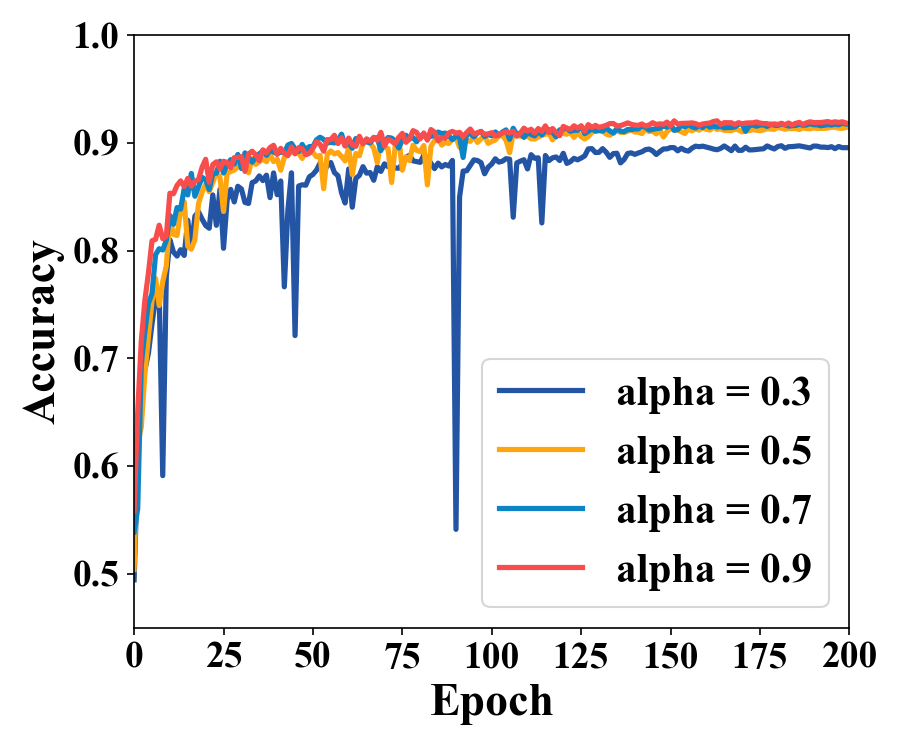}
\end{minipage}
\begin{minipage}{0.49\linewidth}
    \centering
    \includegraphics[width=\linewidth]{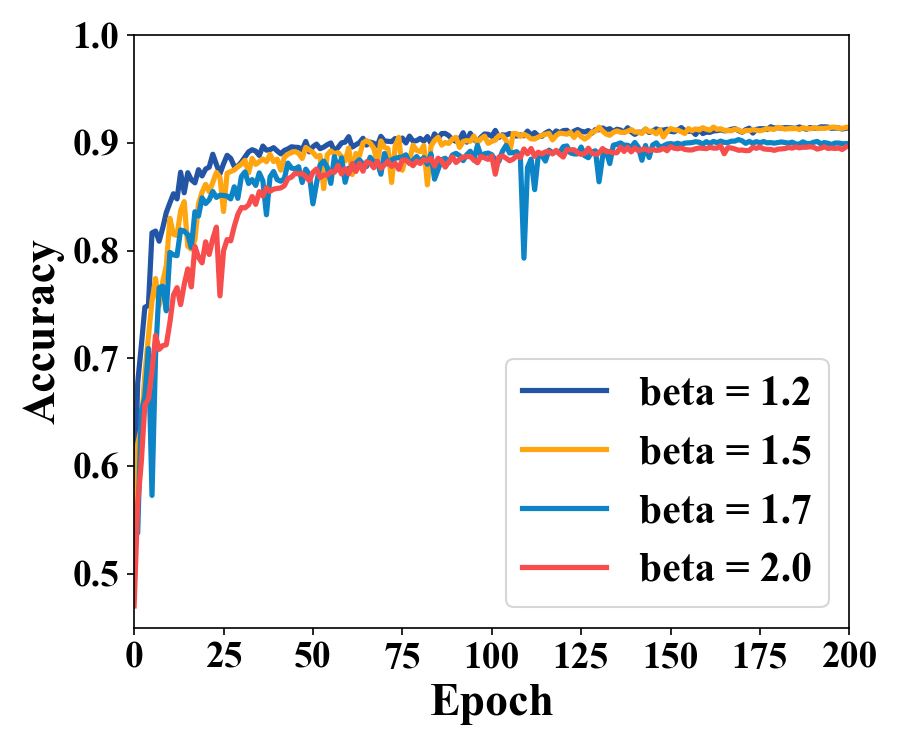}
\end{minipage}
\vspace{-12pt}
\caption{The effect of $\alpha$ and $\beta$ on modular training.}
\label{fig:rq4_alpha_beta}
\vspace{-10pt}
\end{figure}

In this research question, we investigate the effect of major parameters on \projectNameCNN, including $\alpha$ and $\beta$ (the weighting factors in Eq. \ref{eq:objective}), as well as $\tau$ (the threshold value, as described in Sec. \ref{subsec:modularizing}). 
Due to the page limit, we only present and discuss the results of ResNet18 on CIFAR10. Nonetheless, we conducted experiments on the remaining three models, and the results are similar, as detailed on the project webpage~\cite{mwt}.

The values of $\alpha$ and $\beta$ will directly influence the modular training process.
The left three sub-figures in Figure \ref{fig:rq4_alpha_beta} illustrate the modular training convergence of coupling loss, cohesion loss, and accuracy on ResNet18-CIFAR10 with the values of $\alpha$ as 0.3, 0.5, 0.7 and 0.9.
After 200 epochs,
as the value of $\alpha$ increases, the cohesion loss gradually decreases.
Moreover, coupling loss and accuracy exhibit insensitivity to changes in $\alpha$ when the value of $\alpha$ lies within a reasonable range (e.g., 0.5 to 0.9). 
Nonetheless, too small values of $\alpha$ may affect the modular training process. 
For instance, when $\alpha$=0.3, the coupling loss is greater than that of $\alpha$=0.9, and the oscillation of accuracy is more prominent.
Additionally, the right three sub-figures in Figure \ref{fig:rq4_alpha_beta} depict the convergence of the three metrics with the values of $\beta$ as 1.2, 1.5, 1.7 and 2.0.
We observe that changes in $\beta$ do not significantly affect coupling loss and accuracy.
Nevertheless, cohesion loss increases notably with an increase in $\beta$.
Overall, under different parameter settings, the performance of \projectName is predictable within certain ranges, as the coupling loss and accuracy are not affected significantly, and the trend of change in cohesion loss is obvious.
The results also show that our default settings (i.e., $\alpha{=}0.5$ and $\beta{=}1.5$) are appropriate.

The value of $\tau$ directly affects the results of modularizing and module reuse. As the threshold increases from 0.1 to 0.9, the KRR of the modules gradually decreases, from 37.36\% to 24.74\%, the cohesion increases from 0.8572 to 0.9437, and the coupling decreases from 0.3594 to 0.2412. Regarding the effect on module reuse, as the threshold increases, the KRR of the module decreases, from 72.57\% to 50.51\%. Nonetheless, the decrease in KRR has a negligible impact on the accuracy of the module, which only drops from 97.77\% to 97.23\%. The details are available on the project webpage~\cite{mwt}.

\begin{tcolorbox}[left=2pt,right=2pt,top=2pt,bottom=2pt]
The performance of \projectName is predictable within certain ranges
in terms of cohesion, coupling, and accuracy under different parameter settings, making it easy to configure to various models.
\end{tcolorbox}

\section{Discussion}
\subsection{The Generality of \projectName}
We argue that \projectName is generalizable to diverse DNN models for the following reasons. First, the calculation of the coupling and cohesion based on sets of relevant weights is general for DNN models. Sets of weights can be constructed at the granularity of individual weights for any neural network, and at the granularity of sub-structures for neural networks with sub-structures. Regarding the recognition of relevant weights, existing works have explored methods such as neuron activations~\cite{fse2020modularity} and weight coverage frequency~\cite{ReMos} to identify relevant weights for different types of neural network models. Consequently, sets of relevant weights can be constructed for any neural network model to evaluate cohesion and coupling. In addition, modular training adds cohesion and coupling loss functions on top of standard training, which is a common way to improve the performance of models in certain aspects by adding new loss functions.

More specifically, a concrete approach generalizable for DNN models could be designed at the granularity of individual weights. The elements in a mask represent the relevance of the corresponding weights to a class. The feasibility of this idea could be supported to some extent by the existing work~\cite{icse2023}.
Moreover, for neural networks with substructures, a specialized approach could be designed at the granularity of substructures. In addition to CNNs, taking Transformers as an example, the elements of a mask can represent the relevance of corresponding attention heads and word embeddings to a class. Some related work~\cite{shi2022compressing,michel2019sixteen} could support the feasibility of the idea to some extent.

\subsection{Threats to Validity}
\textbf{External validity:}
Threats to external validity relate to the generalizability of our results. 
In this paper, we have only evaluated \projectName on CNNs, and the effectiveness on other types of DNNs, such as LSTM and Transformer, remains to be evaluated. However, as discussed above, \projectName is considered to be general, and we will further investigate it in our future work.
Moreover, \projectNameCNN is not validated on larger scale datasets such as ImageNet~\cite{imagenet}, due to the huge resource and time consumption for training models on large-scale datasets. 
Additionally, our experiments did not consider situations where the original models are not highly accurate.
We leave these as future work.

\textbf{Internal validity:}
An internal threat comes from the choice of models and datasets. To mitigate this threat, we use CIFAR-10 and SVHN datasets as well as VGG16 and ResNet18 models from PyTorch~\cite{paszke2019pytorch}, which are well organized and widely used.

\textbf{Construct validity:}
A threat relates to the suitability of our evaluation metrics. 
The concepts of cohesion and coupling in the context of DNN modularization are first proposed in this paper, thus evaluating cohesion and coupling of DNN models remains an open problem. Other metrics for calculating overlap may also be suitable for measuring cohesion and coupling; however, Jaccard Index is a representative metric and has been widely used in existing work to measure differences between modules~\cite{qi2022patching,fse2020modularity,nnmodularity2022icse}%

\section{Related Work}

\subsection{DNN Modularization}
Existing DNN modularization studies~\cite{qi2022patching,nnmodularity2022icse,fse2020modularity}  and other related efforts~\cite{ReMos,icse2023,zhang2020dynamic} identify neurons or weights in the pre-trained model that is relevant to the target class and retain only these relevant neurons or weights to construct a module responsible for the target class. These works can be classified into two categories based on their identification of relevant neurons and weights: \textit{neuron activation-based}~\cite{nnmodularity2022icse,fse2020modularity,ReMos,zhang2020dynamic} and \textit{search-based}~\cite{qi2022patching,icse2023} modularization approaches. 
The neuron activation-based approach determines whether a neuron is activated by the samples belonging to the target class based on whether its output is greater than zero, and further measures the relevance of the neuron or its associated weights to the target class. For instance, ReMos~\cite{ReMos} measures relevance of weights to the target class mainly by computing neuron coverage frequency~\cite{deepxplore} and weight coverage frequency, and the weights with relevance above a threshold are considered relevant. 

The search-based approach measures the relevance of weights retained in a candidate module to the target class based on the recognition ability of the candidate for the target class and the candidate's size or difference. For instance, CNNSplitter~\cite{qi2022patching} employs genetic algorithm to search for relevant convolution kernels. It assesses the relevance of kernels in a candidate module by evaluating the classification accuracy of the candidate for the target class and measuring the difference between candidates with Jaccard distance. The weights contained in the resulting candidate with the highest classification accuracy and difference are considered relevant.

Unlike all existing works, modularizing while training proposed in this paper is a new paradigm. By considering the cohesion and coupling in the training process, \projectName can overcome the limitations of the paradigm of modularizing after training and achieve more efficient and effective modularization than existing works.

\subsection{DNN Pruning}
DNN pruning techniques, such as iterative magnitude pruning~\cite{pruning_hansong, lottery, rosenfeld2021predictability}, remove some of the weights that are not important for the whole task to generate a smaller model, thus reducing the resources and time required for inference on the whole task. In contrast, our work removes some of the convolution kernels that are irrelevant to the sub-task to decompose the model into modules, thus facilitating module reuse. The above difference in objectives further leads to a difference in loss functions. The loss functions for DNN pruning techniques usually involve the number of weights in the whole model and the magnitude of the weights. However, in our work, the cohesion loss and coupling loss functions involve the overlap between the weights corresponding to different samples.

\section{Conclusion}
In this work, we propose \projectName, a novel paradigm for realizing modularization of DNN models to improve their reusability. 
To overcome the limitations of the existing methods that achieve modularization on trained models, we take a different approach that incorporates the modularization into the model training process.
In doing so, our \projectName integrates the cohesion loss and the coupling loss with the normal cross-entropy training loss, which drives the training process to optimize the modular characteristics as well as the accuracy at the same time. Specifically, we implement \projectName on CNN models. The experimental results demonstrate that \projectName substantially reduces the size of the resultant modules and the time cost of modularization while incurring less performance loss than the state-of-the-art approach~\cite{qi2022patching}. For instance, the kernel retention rate of modules generated by \projectName is only 14.58\%, with a reduction of 74.31\% over the state of the art. Furthermore, the time cost required for training and modularizing is 108 minutes, half of the time required by the state-of-the-art approach.

In the future, we plan to extend MwT to other types of DNNs, such as LSTM and Transformer. Additionally, we will explore various scenarios of model reuse, including the reuse of modules from different pretrained models.

\textbf{Data Availability}: Our source code and experimental data are available at: \textbf{\url{https://github.com/qibinhang/MwT}}.

\begin{acks}
This work was supported partly by National Natural Science Foundation of China Under Grant Nos (61972013 and 61972013) and partly by Guangxi Collaborative Innovation Center of Multi-source Information Integration and Intelligent Processing.
\end{acks}

\balance
\bibliographystyle{ACM-Reference-Format}
\bibliography{reference}

\end{document}